\documentclass[journal]{IEEEtran}
\usepackage[colorlinks,linkcolor=black,anchorcolor=black, citecolor=black,]{hyperref}

\usepackage{graphicx}
\usepackage[ruled,vlined]{algorithm2e}
\usepackage{bm}
\usepackage{subfigure}
\usepackage{amsmath}
\usepackage{amsfonts}

\usepackage{multirow}

\ifCLASSINFOpdf
\else
\fi
\begin{document}
%
\title{A Bayesian Federated Learning Framework with Online Laplace Approximation}
%
%
%
\author{Liangxi Liu$^*$, Xi Jiang$^*$, Feng Zheng$^\dagger$,~\IEEEmembership{Member,~IEEE}, Hong Chen, Guo-Jun Qi,~\IEEEmembership{Fellow,~IEEE}, Heng Huang, Ling Shao,~\IEEEmembership{Fellow,~IEEE}   %
    \thanks{Manuscript received 22 Jul. 2021; revised 9 Aug. 2023; accepted 23 Sep. 2023. This work was supported by the National Key R\&D Program of China (Grant NO. 2022YFF1202903) and the National Natural Science Foundation of China (Grant NO. 62122035). Recommended for acceptance by M. Sugiyama.  ($^*$\emph{Co-first authors: Liangxi Liu and Xi Jiang, }$^\dagger$\emph{Corresponding author: Feng Zheng})}%
    \thanks{L. Liu, X. Jiang, and F. Zheng are with Southern University of Science and Technology, Shenzhen 518055, China (email: liulx@mail.sustech.edu.cn, jiangx2020@mail.sustech.edu.cn, and f.zheng@ieee.org).}
    \thanks{H. Chen is with the College of Informatics, Huazhong Agricultural University, Wuhan 430070, China, and also with the Engineering Research Center of Intelligent Technology for Agriculture, Ministry of Education, Wuhan 430070, China. (email: chenh@mail.hzau.edu.cn)}
    \thanks{G. Qi is with Westlake University, Bellevue, WA 98006, USA, and also with OPPO Research, Hangzhou, Zhejiang 310030, China. (e-mail: guojunq@gmail.com)}
    \thanks{H. Huang is with the University of Maryland College Park, College Park, MD, USA. (email: heng.huang@pitt.edu)}
    \thanks{L. Shao is with the UCAS-Terminus AI Lab, University of Chinese Academy of Sciences, Beijing 100049, China (e-mail: ling.shao@ieee.org).}
    \thanks{Digital Object Identifier 10.1109/TPAMI.2023.3322743}

    \thanks{Code is available at \href{https://github.com/Klitter/A-Bayesian-Federated-Learning-Framework-with-Online-Laplace-Approximation}{{\color{black}https://github.com/Klitter/A-Bayesian-Federated-Learning-Framework-with-Online-Laplace-Approximation}}}
    } 

%
%

\markboth{}%
{}
%



\maketitle

\begin{abstract}
 Federated learning (FL) allows multiple clients to collaboratively learn a globally shared model through cycles of model aggregation and local model training, without the need to share data. 
 Most existing FL methods train local models separately on different clients, and then simply average their parameters to obtain a centralized model on the server side. However, these approaches generally suffer from large aggregation errors and severe local forgetting, which are particularly bad in heterogeneous data settings. 
 To tackle these issues, in this paper, we propose a novel FL framework that uses online Laplace approximation to approximate posteriors on both the client and server side. On the server side, a multivariate Gaussian product mechanism is employed to construct and maximize a global posterior, largely reducing the aggregation errors induced by large discrepancies between local models. 
 On the client side, a prior loss that uses the global posterior probabilistic parameters delivered from the server is designed to guide the local training. 
 Binding such learning constraints from other clients enables our method to mitigate local forgetting. 
 Finally, we achieve state-of-the-art results on several benchmarks, clearly demonstrating the advantages of the proposed method. 
\end{abstract}


%
\IEEEpeerreviewmaketitle

\section{Introduction}\label{SEC:Introduction}
%
%
%
%

Traditional machine learning requires data to be aggregated in a centralized manner. However, due to potential privacy leaks and communication overheads, it is unrealistic to centralize data when there are multiple parties participating in the learning process.
Thus, FL (FL) has been introduced to train a globally shared model without the need to transfer data from multiple parties \cite{yang2019federated}. 
Due to the strict requirements to protect data privacy, each participant can only access its own data to train a local model and optimize a local objective function.
Thus, instead of directly optimizing a global objective function over all data, FL aims to optimize an overall separable objective function constructed from the sum of all local objective functions of participants in a data-isolated environment.

In practice, the standard FL paradigm involves two iterative stages:
(1) Aggregation: A global model $M_{S}$ is aggregated in the server using local models from clients and then distributed to these clients.
(2) Local Training: After receiving $M_{S}$, each client trains its model on its own dataset $D_{n}$, and then uploads its local model to the server.
After a few series of iterations, FL can obtain a global model that achieves similar performance to the traditional centralized approach.
FedAvg \cite{mcmahan2016communication} is the first to propose this paradigm, achieving competitive performance to several centralized approaches on homogeneous data (i.e. data that has an identical statistical distribution for different clients).

\begin{figure}[t]
\centering
\includegraphics[width=9cm]{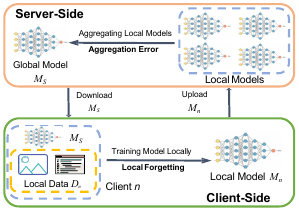}
\caption{Illustration of iterative steps and two problems of federated learning, i.e., aggregation error and local training. In this paper, the two problems arise during the aggregation process on server and the local training stages on clients.}
\label{Regression}
\end{figure}

However, since real-world environments typically differ among clients, local data from different clients generally follow to different statistical distributions, i.e. are heterogeneous.
As shown in \cite{li2019convergence,zhao2018federated}, compared to the results on homogeneous data, both the convergence rate and final accuracy of FedAvg \cite{mcmahan2016communication} on heterogeneous data distribution are significantly reduced.
The inherent reason for this performance degradation is the fact that the heterogeneity of data over clients makes locally optimized neural networks heterogeneous as well, resulting in a multimodal mixture of local posteriors.
Further, the simple strategy applied in FedAvg is actually equivalent to optimizing models on a mixture of local posteriors (discussed in Sec. \ref{SEC:Method}).
Such a multimodal mixture inevitably leads to two problems: large aggregation error and severe local forgetting, which are discussed in more detail as follows.



\begin{figure*}[t]
    \centering
    \subfigure[]{
        \includegraphics[width=3in]{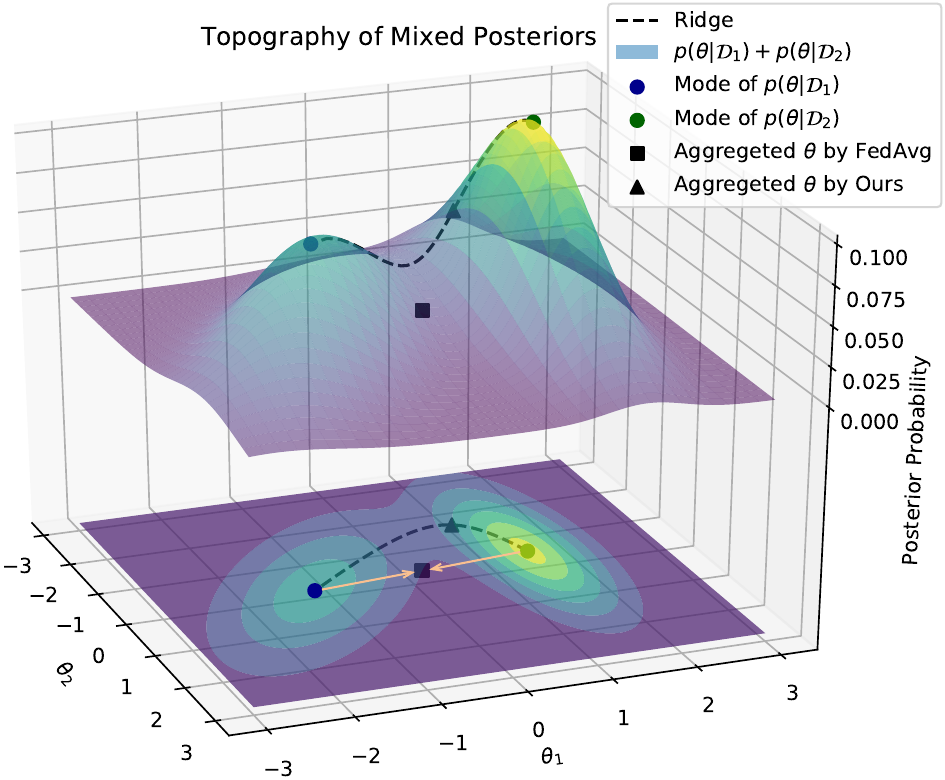}
        \label{AELF3D_1}
    }
    \subfigure[]{
	\includegraphics[width=3in]{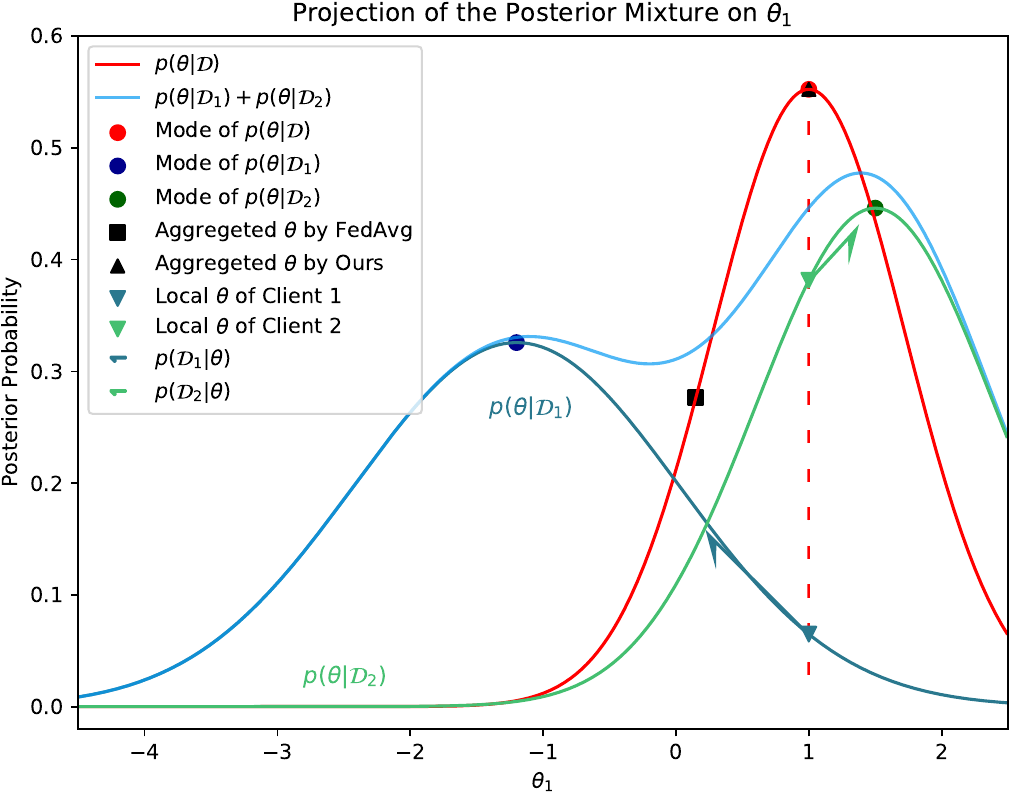}
        \label{AELF3D_2}
    }
    \caption{
   The problems of AE and LF are illustrated on two clients.
    (a) Mixture of two heterogeneous Gaussians $p(\theta | \mathcal{D}_{1})$ and  $p(\theta | \mathcal{D}_{2})$ with bivariance $\theta=[\theta_{1}, \theta_{2}]$ in a three-dimensional space.
    (b) Corresponding projection of the posterior mixtures on the $\theta_{1}$ axis.
    The lines and shapes represent the following.
    In both figures, the blue and green dots are the corresponding modes of $p(\theta | \mathcal{D}_{1})$ and  $p(\theta | \mathcal{D}_{2})$,
    and the triangle and square are the aggregated results of our method and FedAvg, respectively.
    In the left figure, the dashed line is the ridge of a mixed density, consisting of two modes, anti-modes and saddle points.
    In the right figure, the blue and green solid lines represent local posterior $p(\theta | \mathcal{D}_{1})$ on client $1$ and $p(\theta | \mathcal{D}_{2})$ on client $2$.
    Additionally, the global posterior $p(\theta | \mathcal{D})=p(\theta | \mathcal{D}_{1}, \mathcal{D}_{2})$ is represented as a solid red line, and the dashed red line maps the global mode (a red dot) into local parameters, which are blue and green inverted triangles on each client, respectively.
    Intuitively, we can see that the aggregated result of the local modes for FedAvg lies in the valley of the mixture, while our result lies on the ridge. 
    Meanwhile, the figure on the right illustrates that our aggregated result is a global mode, while that of FedAvg is located at the halfway point of the global posterior.  
    Thus, we can see that a large gap exists between the global optimal model and the results of aggregating the local models through FedAvg. 
    The two are more closely aligned for our method.
    Besides, as shown in the figure on the right, the local likelihood function for the local training will drag the local parameters to the corresponding local modes, which deviate from the heterogeneous modes.
    This is our concern; that local training makes models forget knowledge learned from other heterogeneous clients.
}
    \label{FIGAELF}
\end{figure*}

\textbf{Aggregation error - from server side aggregation:}
A standard aggregation method for FL is proposed in FedAvg \cite{mcmahan2016communication}, which simply takes a weighted element-wise average on the parameters of local models using the sample ratios of the clients. 
This simple averaging method is effective in learning over homogeneous data, since the parameters of local models share exactly the same posterior
probability distribution, which is also equal to the global posterior of the server. 
However, for heterogeneous data distribution, parameters of different local models have diverse posterior probability distributions, as shown in Fig. \ref{FIGAELF}.
Due to the discrepancy between the local modes, simply averaging the parameters will cause the aggregated posterior probability to have a larger uncertainty than that of the homogeneous data.
For example, when the posterior follows a Gaussian distribution, the classical FedAvg is equivalent to collapsing a mixture of local Gaussian posteriors into a Gaussian.
Thus, it always yields a model with low posterior probability because of the unreasonable multimodal mixture of local heterogeneous posteriors, as shown in Fig. \ref{FIGAELF}.
Therefore, the model aggregated on the server undoubtedly lacks confidence in predictions and, consequently, the degree of generalization is also low.
For simplicity, in this paper, we refer to this problem as aggregation error (AE). 


\textbf{Local forgetting - from client side training:}
After the server side aggregation, the global model is distributed to multiple clients as an initialization for further local training.
When training over homogeneous data, local likelihood distributions are identical for all clients. As such, the locally optimized models are able to generalize well among clients by maximizing the local likelihood distributions.
However, the generalization ability of local models is significantly reduced when training over heterogeneous data distribution.
After a globally optimized model is received from the server side and trained on the client side, local parameters are only equal to the global mode at the beginning of training.
During local training, because of the heterogeneous likelihoods, the original local parameters are shifted and pushed to the local modes, as demonstrated in Fig. \ref{FIGAELF}.
This causes the local probabilities to be pushed far away from the global posterior, as well as other local heterogeneous posteriors.
Once the local training is completed, the local models can perform well on the corresponding local dataset, but usually achieve low precision on heterogeneous clients.
In other words, local training makes local models forget the knowledge learned from other clients.
%
Such forgetting will lead to large AE in the next round of training due to the displacement of local modes.
In this paper, we refer to this reduction in the ability to generalize the local models to heterogeneous data as local forgetting (LF). 



To tackle the issues mentioned above, we propose a novel FL framework from a Bayesian perspective.
The most direct benefit of this probabilistic framework is that we can use Gaussian distributions to approximate both the local posterior of each client and the global posterior of the server. 
Thus, the original problem of probabilistic optimization can be easily transformed to a Laplace approximation problem, for which several established properties and algorithms can be used.
Through the variational inference of posteriors, we can theoretically analyze why the parameter averaging induces the problems of AE and LF, and successfully solve them by maximizing the global posterior in both the aggregation and local training steps.

On the server side, we apply the Gaussian product method to obtain the expectation and covariance of the global posterior probability by multiplying the local posteriors with the local posterior probabilistic parameters uploaded by the clients.
The product of a Gaussian distribution is also strictly a Gaussian form, for which a convex function can be easily optimized.
Conversely, the mixture of Gaussian used for parameter averaging may not necessarily be Gaussian.
Additionally, the previous work \cite{ray2005topography} analyses the topography of multivariate Gaussian mixtures and proves that the model parameters obtained by the Gaussian product lie on the ridges. 
Accordingly, our aggregation method can obtain a better posterior probability in a product of posteriors than the simple averaging method, thus reducing the aggregation error.

On the client side, we develop a prior iteration (PI) strategy, treating the global posterior probabilistic parameters distributed from the server as priors.
With PI, we derive a prior loss from the prior distribution for local training. 
Minimizing the prior loss is equivalent to maximizing the global posterior approximated by the product of multiple local posteriors in the last round of training. 
Therefore, by adjusting the weight of both the likelihood and the prior, our method is able to reach a compromise between localization and generalization, thereby maintaining a strong generalization ability by mitigating the local forgetting.

Besides, to effectively approximate a local posterior by a Gaussian in FL, we design a new federated online Laplace approximation (FOLA) module.
By regarding the global posterior probabilistic parameters, including the expectations and covariances delivered from the server, as priors, FOLA can integrate all covariances of the previous FL process in an online manner. 
As a result, FOLA allows us to obtain local posterior probabilistic parameters evaluated online, which can be directly used in an FL framework.

Our contributions can be briefly summarized as follows:
\begin{itemize}
\item 
We first analyze the problems of federated learning from the perspective of posteriors rather than from the perspective of optimization as usual, and attribute them into the problems of aggregation error and local forgetting. 
\item 
Then, from the Bayesian perspective, we propose a novel Federated Online Laplace Approximation (FOLA) method to efficiently approximate Gaussian posteriors in a federated manner, instead of using centralized Laplace Approximations.

\item 
Based on FOLA, we propose a Gaussian product method to construct a global posterior on the server side and a prior iteration strategy to update the local posteriors on client sides, both of which are easy to optimize.
By successfully maximizing these posteriors of the server and clients, we can simultaneously reduce the aggregation error and local forgetting.


\item 
Finally, we conduct experiments on several commonly used FL benchmarks and demonstrate the superiority of our framework compared with strong baselines in terms of various metrics. 
\end{itemize}

The rest of the paper is structured as follows. Section \ref{SEC:RW} describes related work, Section \ref{SEC:Method} details the problem setup and introduces our FL framework including the aggregation strategy and the prior loss. Finally, we report and analyze the experimental results in Section \ref{SEC:EX}, and the draw our conclusions in Section \ref{SEC:Con}.

\section{Related Work\label{SEC:RW}}

\subsection{Bayesian Approximation}
In this section, we first introduce some mathematical foundations to provide strong theoretical support for our model aggregation and local training strategies. 

Based on Bayesian theory, \cite{mackay1992practical,ritter2018scalable} introduce a practical Laplace approximation method to approximate the posterior probability using a Gaussian distribution $\theta \sim \mathcal{N}\left(\theta^{*}, \bar{H}^{-1}\right)$.
Generally, the expectation is set to the optimal parameter $\theta^{*}$, and the positive semi-definite (PSD) diagonal precision is set to the inverse of the average Hessian matrix.
Specifically, if $(x, y)$ is a sample pair with input $x$ and target $y$, the average Hessian matrix can be calculated by $\bar{H} = -\frac{1}{|\mathcal{D}|} \sum_{(x, y) \in \mathcal{D}} H_{\ln p(y \mid x,\theta)}$, where
$H_{\ln p(y \mid x,\theta)}$ is the Hessian of the log posterior $\ln p(y \mid x,\theta)$ for each sample pair $(x,y) \in \mathcal{D}$.
However, both the time and space complexities of directly computing $\bar{H}$ are $O(d^{2})$, where $d=|\theta|$ is the number of parameters, making the above approximation difficult to implement.

In order to reduce the computational overhead, some methods \cite{foresee1997gauss,schraudolph2002fast} use a generalized Gauss-Newton (GGN) matrix $\mathrm{G}$ calculated by the Levenberg-Marquardt algorithm in $O(d)$ to approximate $\bar{H}$. 
The matrix $G$ is defined as $G=\frac{1}{|\mathcal{D}|} \sum_{(x, y) \in \mathcal{D}} J_{f}^{\top} H_{{L}} J_{f}$, where $H_{{L}}$ is the Hessian of the loss ${L}(y, z)$ evaluated on the output $z = f(x, \theta)$, and $J_{f}$ is the Jacobian of $f(x, \theta)$ w.r.t. the parameters $\theta$.
In \cite{amari2012differential,amari1998natural}, a quadratic form of the Fisher information matrix $F=\mathbb{E}_{p(y \mid x,\theta)}\left[\nabla \log p(y \mid x,\theta) \nabla \log p(y \mid x,\theta)^{\top}\right]$ 
is provided.
Thus, the negative expected Hessian of the log likelihood is equal to the Fisher information matrix $-\mathbb{E}_{p(y \mid x,\theta)}\left[H_{\ln p(y \mid x,\theta)}\right]=F$, so the matrix $F$ can be used to approximate $\bar{H}$. 
Some works \cite{park2000adaptive,pascanu2013revisiting} show that the GGN and Fisher matrix are equivalent to each other for several common loss functions, such as the cross-entropy loss and squared loss.

Additionally, the method in \cite{martens2010deep} selects an empirical Fisher information matrix calculated by $\bar{F} = \frac{1}{|\mathcal{D}|} \sum_{(x, y) \in \mathcal{D}} \nabla \log p(y \mid x,\theta) \nabla \log p(y \mid x,\theta)^{\top}$, which is a crude and biased approximation of $F$. 
As mentioned in \cite{martens2016second}, $\bar{F}$ is low-rank so its diagonal can be computed by $diag(\bar{F}) = \frac{1}{|\mathcal{D}|} \sum_{(x, y) \in \mathcal{D}} sq(\nabla \log p(y \mid x,\theta))$, where $sq(\cdot)$ denotes the coordinate-wise square function. 
If we assume $\bar{F}$ is diagonal, both the time and space complexity of calculating $diag(\bar{F})$ are $O(d)$, making the Laplace approximation method easy to implement.
In \cite{lecun1990optimal,lee2017overcoming,ritter2018scalable,zenke2017continual}, the inverse of the diagonal of the empirical Fisher information matrix and the Laplace approximation are combined to approximate the expectation and covariance of the posterior probability, which works well for their target tasks.  
Moreover, in \cite{zenke2017continual}, a derivation of ordinary differential equations for optimization is used to prove that the accumulation of gradient squares multiplied by the learning rate of the optimization steps is equivalent to the diagonal of the Hessian matrix.

Besides, a multivariate normal mixture method \cite{ray2005topography} illustrates that, by using a ridge line manifold that contains all critical points, the topography or density of mixing multiple Gaussian distributions can be analyzed rigorously in lower dimensions. 
Further, all critical points (modes, antimodes and saddlepoints) of $N$-component multivariate normal densitiy are shown to be the points in an $N-1$ dimensional hypersurface $\{ \theta | \theta = (\sum_{n=1}^{N} \pi_{n} \Sigma_{n}^{-1})^{-1}(\sum_{n=1}^{N} \pi_{n} \Sigma_{n}^{-1} \mu_{n}), \alpha \in [0,1] , \sum_{n=1}^{N}\pi_{n} = 1 \}$, where $\mu_{n}$ and $\Sigma_{n}$ are the expectation and covariance of a multivariate Gaussian distribution.

\subsection{Federated Learning} 

Recently, in order to solve collaborative training tasks, the machine learning community has been paying more attention to federated learning. 
FedSGD \cite{mcmahan2016communication}, a method directly adapted from the native SGD, is proposed to update the model on a server by averaging local one-step gradient descents. 
However, this approach has high communication costs during the whole training process until convergence and, more importantly, it leaks information about local data from the gradients \cite{zhu2019deep}.
To overcome these two disadvantages, the FedAvg method in \cite{mcmahan2016communication}, inspired by parallel SGD-related algorithms \cite{zhang2015deep,shamir2014communication,reddi2016aide,zhou2017convergence,stich2018local}, replaces the one-step gradient descent scheme with multiple steps, empirically outperforming FedSGD in both efficiency and accuracy. However, it fails to work well under heterogeneous data settings.

The work in \cite{zhao2018federated} shows that the accuracy reduction caused by heterogeneous data distribution can be explained by weight divergence. 
By creating a set of globally shared samples, the proposed method has an improved accuracy on heterogeneous data distribution but requires extra memory and additional training time on the client side.
Based on FedAvg, FedProx \cite{li2018federated} employs an isotropic penalty term $\frac{\lambda}{2}||\theta-\theta_{s}||$, which restricts local models $\theta$ to be close to the server model $\theta_{s}$. However, the convergence rate of the isotropic FedProx is slowed down in some situations \cite{shoham2019overcoming}.
FedCurv \cite{shoham2019overcoming} improves FedProx by treating the diagonal entries of the Fisher matrix $F$ as the anisotropic stiffness of parameters. 
It mitigates the weight divergence by optimizing a decomposed global posterior $p(\theta | \mathcal{D} ) = p(\mathcal{D}_{n} | \theta) + p(\theta | \bar{\mathcal{D}}_{n} )$ during local training, where $\bar{\mathcal{D}}_{n}$ denotes a complement to the local data $\mathcal{D}_{n}$ in the $n$-th client. 
However, FedCurv is unable to directly reduce the AE problem and, more importantly, it approximates $p(\theta | \bar{\mathcal{D}}_{n} )$ using an offline Fisher information matrix, resulting in a biased evaluation.
Moreover, by taking advantage of the Beta-Bernoulli process, several methods \cite{yurochkin2019bayesian,wang2020federated} employ a novel non-parametric algorithm based on FedAvg for federated optimization. These models solve the problem of the permutation-invariant nature of the neural network by finding the permutation of the parameters before averaging them.


Some recent FL research handles the heterogeneity of local data. FedNova~\cite{wang2020tackling} proposes a scalable aggregation method to consider the different dataset sizes of clients. 
SCAFFOLD~\cite{karimireddy2020scaffold} utilizes control variate to estimate the drift of directions of optimization. The above two methods improve FedAvg in AE and LF separately.  
The non-IID problem can be analyzed more clearly based on the Bayesian approach. 
VIRTUAL~\cite{corinzia2021variational} uses a Bayesian network as the posterior distribution in client optimization. However, VIRTUAL decomposes the variational inference process for multi-task learning but ignores the AE problem, which aggregates received updates with simple probability multiplying, leading to slow convergence. 
FedPA~\cite{al2020federated} also regards each model as a Gaussian and uses Monte Carlo sampling to sample local models during local training and then statistics variance from the samples. It is inefficient because it requires sampling a lot of local historical models. 
FedBE~\cite{chen2020fedbe} also implements Bayesian inference, leveraging the Monte Carlo method. FedBE approximates the model posterior using a Gaussian distribution and constructs a distillation-based method for training the global model. However, it requires additional unlabeled data on the server.  
FedSparse~\cite{louizos2021expectation} uses a Gaussian prior over parameters and maintains a Bernoulli distribution to allow for sparsity in the local parameters. It mainly tackles the challenge of communication costs but has limited performance in model accuracy. In this paper, we intend to solve both AE and LF in a Bayesian variational inference way. 

\section{The Proposed Method\label{SEC:Method}}

In this chapter, we first formalize the federated learning framework and set out the objectives of the maximum posterior estimation. We then analyze the issues that arise from the federation aggregation model, pointing out from the Bayesian posterior perspective that simple aggregation strategies can lead to aggregation errors when client data is heterogeneous. This is because the means of the local posterior Gaussians of different clients can vary, as demonstrated in Fig. \ref{FIGAELF}. To reduce aggregation errors, we propose a method that approximates the global posterior by using the product of posterior probabilities. Simultaneously, in order to minimize local forgetting, we propose a Prior Iteration method to constrain the training objectives of the clients. Finally, to satisfy the efficiency requirements of federated learning, we propose using the Federated Online Laplace Approximation to generate parameters for the local posterior distribution. We also analyze the algorithm complexity and security in the last section.

FL methods aim to obtain a globally optimal model on a central server from the models of multiple clients, which are responsible for collecting data in a privacy-protected manner. 
In fact, to find the optimal parameters $\theta$ in normal settings, maximizing the global posterior $p(\theta|\mathcal{D})$ is a common objective and easy to be solved.
However, in FL, it is impossible to directly evaluate the global posterior $p(\theta|\mathcal{D})$.
Note that $\theta$ denotes the model weights, $\mathcal{D} = \{ \mathcal{D}_{1}, \mathcal{D}_{2},...,\mathcal{D}_{N} \} $ is a set of all data over the clients, and $N$ denotes the number of clients.
The reason why it is difficult to directly estimate the posterior $p(\theta|\mathcal{D})$ lies in that participants are all banned from accessing the data $\mathcal{D}$ in a centralized manner. 
Thus, in order to maximize $p(\theta|\mathcal{D})$, FL methods try to minimize an alternative separable objective function over clients,
\begin{equation}
   \min_{\theta} \mathcal{J}(\theta) = \sum_{n=1}^{N} \pi_{n} \mathcal{J}_{n}(\theta),
\label{FLLOSS}
\end{equation}
where $\pi_n \in [0,1]$ and $\sum_{n=1}^{N}\pi_{n} = 1$. Normally, $\pi_{n}$ is set to $\frac{m_{n}}{m}$, where $m_{n}$ is the number of samples provided by the corresponding client and $m$ is the total number of samples over all clients $m=\sum_{n}^{N} m_{n}$. 
In addition, if we assume that $(x, y)$ is a sample pair with input $x$ and target $y$, $\mathcal{J}(\theta)$ is an overall objective function for the whole learning system and $\mathcal{J}_{n}(\theta) = \mathbb{E}_{(x,y)\sim \mathcal{D}_{n}} [\mathcal{L}(\theta;(x,y))]$ is a local objective function on one client with its own local data $\mathcal{D}_{n}$.

\subsection{Problem Analysis of Aggregation\label{SEC:ProblemAnalyse}}
Generally, minimizing the objectives on clients $\mathcal{J}_{n}(\theta)$ in Eq. \ref{FLLOSS} is equivalent to maximizing the corresponding posterior probabilities.
The relationship between the overall objective function and multiple local posterior probabilities can be derived as:
\begin{equation}
  \min_{\theta} \mathcal{J}(\theta) = \max_{\theta} \sum_{n=1}^{N} \pi_{n} p(\theta | \mathcal{D}_{n}). 
\label{EQ:MixtureLocal}
\end{equation}

Actually, by maximizing the above mixture of local posteriors, classical FL methods aim to maximize $p(\theta|\mathcal{D})$ indirectly.
To achieve this, most existing methods approximate the global posterior using a mixture of local posteriors $p(\theta | \mathcal{D}) \approx \sum_{n=1}^{N} \pi_{n} p(\theta | \mathcal{D}_{n})$.
Although the optimal parameter obtained in Eq. \ref{EQ:MixtureLocal} is not necessarily the global optimal parameter of $p(\theta | \mathcal{D})$, it generally works sufficiently well in practice \cite{mcmahan2016communication}.

While the global posterior $p(\theta | \mathcal{D})$ can, to some extent, be well approximated by a mixture of local posteriors, it is difficult to optimize directly because local posteriors are intractable to compute for neural networks. 
Fortunately, in practice, a Gaussian distribution can be used to approximate the posteriors in a small neighborhood of the MAP estimate on single clients~\cite{ corinzia2021variational, huang2021rethinking, nguyen2018variational}. Although the Gaussian approximation introduces some error, on one hand, our client-side training is based on the same Gaussian prior, which will be described in Sec. ~\ref{SEC:PriorLoss}. On the other hand, the error will decrease in the process of model parameters approaching the optimal values through continuous iteration, as the first-order derivative is zero at the optimal value. 
Considering the Laplace approximation under the theoretical framework of Bayesian neural networks in ~\cite{mackay1992practical,ritter2018scalable,huang2021rethinking}, without loss of generality, we assume that each posterior $p(\theta | \mathcal{D}_{n})$ follows a multivariate Gaussian distribution $q_{n}(\theta)$. This assumption is also used in variational continual learning~\cite{nguyen2018variational}, which acknowledges that while the true posterior distribution is certainly far more complex than the approximated distribution, leading to potential loss of information, this can be mitigated by storing historical data in memory. This effectively corresponds to the multi-round communication and training process in federated learning.

Furthermore, since the data of different clients are not entirely identical, different local posterior probabilities will have different local expectations and covariances. 
Thus, the local posterior probabilities can be defined as:
\begin{eqnarray}
   p(\theta|\mathcal{D}_{n}) \approx q_{n}(\theta) \equiv \mathcal N(\theta | \mu_{n} , \Sigma_{n})\label{EQ:LocalPosteriorApprox},
\end{eqnarray}
where $(\mu_{n},\Sigma_{n})$ is a pair made up of the expectation and the covariance of the local posterior $p(\theta|\mathcal{D}_{n})$.
The details on how to estimate the local expectations and covariances on clients will be introduced in Sec. \ref{SEC:Estimation}.

After applying Eq. \ref{EQ:LocalPosteriorApprox} to Eq. \ref{EQ:MixtureLocal}, the mixture of local posteriors can be directly solved by a Gaussian mixture. 
However, finding modes on a Gaussian mixture \cite{carreira2000mode} requires an extra heavy computing process.
One efficient strategy is to collapse the mixture of $q_{n}(\theta)$ into a convex function, such as a Gaussian distribution function. 
Typically, it is easy to find the optima in a Gaussian function.
In fact, previous works \cite{mcmahan2016communication,li2018federated} using the traditional FL framework are equivalent to optimizing $\theta$ on a multivariate Gaussian distribution by collapsing the mixture of $q_{n}(\theta)$. 
Thus, like the above assumption for the local posterior, we also assume that the global posterior is a Gaussian with an expectation $\mu_{S}$ and covariance $\Sigma_{S}$:
\begin{equation}
   p(\theta|\mathcal{D}) \approx q_{S}(\theta) \equiv \mathcal N(\theta | \mu_{S} , \Sigma_{S}). \label{EQ:GlobalPosteriorApprox}
\end{equation}

In order to collapse the mixture of $q_{n}(\theta)$ into $q_S(\theta)$, we minimize the Kullback–Leibler (KL) divergence between them. 
The KL divergence is a well-known difference measure between two probability distributions.
Obviously, minimizing the KL divergence will make the mixture of $q_{n}(\theta)$ closer to $q_S(\theta)$, which is defined as
\begin{equation}
 \mu_{S}, \Sigma_{S} = \min_{\mu_{S}, \Sigma_{S} } KL\left(\sum_{n}^{N} \pi_{n} q_{n}(\theta) \| q_{S}(\theta)\right). \label{EQ:KLM}
\end{equation}

Once the KL divergence between the mixture of $q_{n}(\theta)$ and $q_{S}(\theta)$ is minimized, we are able to obtain the expectation and covariance of $q_{S}$:
\begin{eqnarray}
 \mu_{S}  &=& \sum_{n}^{N} \pi_{n} \mu_{n}, \label{EQ:AvgMu}\\
 \Sigma_{S}  &=&  \sum_{n=1}^{N} \pi_{n}( \Sigma_{n} + \mu_{n} \mu_{n}^{\top} - \mu_{S} \mu_{S}^{\top} ). \label{EQ:Sigma}
\end{eqnarray}

This can be further simplified by  assuming $\Sigma_{S}$ is an identity matrix $I$, which is not involved in the minimization of Eq. \ref{EQ:KLM}.
In traditional FL frameworks, FedAvg aggregates local models using Eq. \ref{EQ:AvgMu}. 
In addition, FedProx improves FedAvg by adding an isotropic penalty term during local training, which is equivalent to treating $\mathcal N(\theta | \mu_{S} , I)$ as the prior in  the local training of the next round. More details are discussed in \ref{SEC:PriorLoss}. 

When the data is homogeneous over clients, all $q_{n}(\theta)$ have a similar expectation and covariance, making their mixture convex-like in shape. 
In this case, the minimum value of the KL divergence obtained in Eq. \ref{EQ:KLM} is very small, resulting in an accurate approximation of $q_{S}(\theta)$ to $\sum_{n}^{N} q_{n}(\theta)$. 
Thus, joint modes over all local posteriors can be easily found by Eq. \ref{EQ:AvgMu} in the traditional FL framework. Further, the displacement of modes during local training is small because the joint modes are close to the local modes. 
In other words, under the homogeneous data setting, the aggregation error of the traditional FL framework is inconspicuous and acceptable.

However, when the data is heterogeneous over clients, the mixture of $q_{n}(\theta)$ is non-convex and multimodal, making the above approximation of Eq. \ref{EQ:KLM} imprecise. 
As discussed in Sec. \ref{SEC:Introduction}, aggregating models using previous methods \cite{mcmahan2016communication,li2018federated} on the multimodal mixture inevitably results in a large aggregation error.

\subsection{Bayesian Aggregation Strategy on Server - Multivariate Gaussian Product} \label{SEC:OurAgg}

In order to reduce the aggregation error, we propose a new constructed approximation of $p(\theta | \mathcal{D})$ using a product of local posteriors instead of the mixture. 
Minimizing the objectives on clients $\mathcal{J}_{n}(\theta)$ in Eq. \ref{FLLOSS} is actually equivalent to maximizing the logarithm of the corresponding posterior:
\begin{eqnarray}
   \min_{\theta} \mathcal{J}(\theta) &=& \min_{\theta} \sum_{n=1}^{N} \pi_{n} \mathcal{J}_{n}(\theta) \\
   &=& \max_{\theta} \sum_{n=1}^{N} \pi_{n} \ln p(\theta | \mathcal{D}_{n}) \\
   &=&  \max_{\theta} \ln \prod_{n=1}^{N} p(\theta | \mathcal{D}_{n})^{\pi_{n}}. \label{EQ:FLPro} 
\end{eqnarray}

Therefore, it is reasonable to approximate the global posterior by the product of all local posteriors $p(\theta | \mathcal{D}) \approx \prod_{n}^{N} p(\theta | \mathcal{D}_{n})^{\pi_{n}}$. 
After taking the above approximations of Eq. \ref{EQ:GlobalPosteriorApprox} and \ref{EQ:LocalPosteriorApprox}, we are able to approximate $q_{s}(\theta)$ as the product of all $q_{n}(\theta)$. 

As we know, the product of Gaussians is still a Gaussian form.
In contrast, a mixture of Gaussians is not necessarily a Gaussian.
Thus, we can derive the following equation for $q_{S}(\theta)$:
\begin{equation}
\begin{aligned}
   q_{S}(\theta) \propto \prod_{n=1}^{N} q_{n}(\theta)^{\pi_{n}}. \label{EQ:ProductGaussian}
\end{aligned}
\end{equation} 

Consequently, the global expectation $\mu_{S}$ and covariance $\Sigma_{S}$ can easily be obtained from Eq. \ref{EQ:ProductGaussian} by aggregating the local expectations $\mu_{n}$ and covariances $\Sigma_{n}$:
\begin{eqnarray}
   \mu_{S} = \Sigma_{S} (\sum_{n=1}^{N} \pi_{n} \Sigma_{n}^{-1} \mu_{n}), \label{SMu} \quad \Sigma_{S}^{-1}  =  \sum_{n=1}^{N} \pi_{n} \Sigma_{n}^{-1}.\label{SSigma}
   \label{EQ:ModePro}
\end{eqnarray}

Unlike previous methods, our method approximates the global posterior as the product of $q_{n}(\theta)$, which is always convex. 
Thus, we can efficiently find modes using Eq. \ref{EQ:ModePro} on this a product directly. 
Without the approximation errors between a non-convex multimodal mixture and a convex function, our aggregation strategy can achieve a smaller aggregation error compared to previous works.

In addition, it is worth noting that the result obtained by Eq. \ref{EQ:ModePro} is the mode of Eq. \ref{EQ:ProductGaussian}, while is located on a critical hypersurface of the mixture of $q_n(\theta)$.
Theorems provided in \cite{ray2005topography} systematically analyze the topography of $N$-component multivariate Gaussian density mixtures. 
They clearly show that, compared to simply averaging, our strategy using the Gaussian product method achieves a larger global posterior probability. 
Moreover, the $N-1$ dimensional hypersurfaces obtained by $\sum_{n}^{N} \beta_{n} q_n(\theta)$ with mixing constant  $\beta_{n} \in [0,1] $ and $\sum_{n=1}^{N} \beta_{n} = 1$, under the conditions of $\pi_{n} \in [0,1] $ and $\sum_{n=1}^{N} \pi_{n} = 1$, are the ridges. There consist of all kinds of critical points such as modes, antimodes and saddlepoints.
Therefore, we can conclude that the parameters obtained by our aggregation method are more likely to be located on the desired ridges, while those obtained by simple averaging may be located in valleys, especially when the mixture of local posteriors has multiple optima. 
As a result, since a relatively high global posterior probability can be approximated by a product, our method is able to aggregate a global model in Eq. \ref{EQ:ModePro} that achieves better results on samples across all the clients than the simple averaging method. 
In other words, the aggregation error is successfully reduced.

\subsection{Bayesian Training Strategy on Clients - Prior Iteration} \label{SEC:PriorLoss}
On the client side, the local posterior $p(\theta|\mathcal{D}_{n})$ can be decomposed into a likelihood $p(\mathcal{D}_{n}|\theta)$ and a prior $p(\theta)$ by Bayes' theorem as follows:
\begin{equation}
\begin{aligned}
   \ln p(\theta|\mathcal{D}_{n}) = - \mathcal{J}_{n}(\theta) = \ln p(\theta) + \ln p(\mathcal{D}_{n}|\theta) - \ln p(\mathcal{D}_{n}),\label{BClient}
\end{aligned}
\end{equation}
where $ p(\mathcal{D}_{n})$ is a constant when the dataset $\mathcal{D}_{n}$ is given. 

In particular, we further propose a novel local training strategy, called prior iteration (PI), which regards the posterior $p^{*} (\theta | \mathcal{D})$ corresponding to the aggregated model distributed from the cloud as a prior $p(\theta) = p^{*} (\theta | \mathcal{D}) \approx \mathcal N(\theta | \mu_{S}^{*}, \Sigma_{S}^{*})$.
In fact, in the traditional FL methods, considering the aggregated model as an initial model for local training \cite{mcmahan2016communication,li2018federated} is equivalent to treating the aggregated model parameters as priori parameters.
Furthermore, in continuous learning, the posteriors of old tasks can also be treated as priors when learning a new task.
Thus, similarly, we can regard the global optimization of the entire data $\mathcal{D}$ as a global task and the local optimization a separate subtask.
In this case, the posterior probability of global parameters on the cloud becomes the prior probability of local parameters on clients.
Therefore, we rewrite the local posteriors as follows:
\begin{equation}
\begin{aligned}
   \ln p(\theta|\mathcal{D}_{n}) = \ln p^{*} (\theta | \mathcal{D}) + \ln p(\mathcal{D}_{n}|\theta) - \ln p(\mathcal{D}_{n}). \label{BClientPI}
\end{aligned}
\end{equation}

This strategy is an indispensable part of our Bayesian federation learning framework because it guarantees that the global covariance obtained by our method can be evaluated in an online manner. The existing similar work, such as FedCurv, evaluate the anisotropic stiffness of parameters in an offline manner.  
The existing similar work, such as FedCurv, evaluates the anisotropic stiffness of parameters in an offline manner. That is to say, after each client training is completed, an extra epoch is needed to obtain the stiffness of the current model parameters, during which $p(\theta | \bar{\mathcal{D}}_{n})$ is used to approximate $p(\theta_s | \bar{\mathcal{D}}_{n})$.
Compared to offline methods, the online method generally works better and the reasons will be detailedly discussed in Sec. \ref{SEC:Estimation}.
Similar to the model parameters, the covariance matrices in our model will be distributed to local clients and aggregated on the server side as well.
In fact, $\Sigma_{S}^{*}$ consists of all local covariance matrices uploaded in the last round of training, and the local covariance obtained in the current round is made up of a Fisher information matrix evaluated and online a priori covariance matrix $\Sigma_{S}^{*}$.
Details on how to aggregate the global covariance and how to evaluate local covariances will be introduced in Sec. \ref{SEC:Estimation}.

Besides, a second-order regularization term can be derived directly from the prior to guide the learning throughout the training process and mitigate local forgetting. Thus, by applying a logarithm to $p(\theta)$, a prior loss considering such a second-order regularization term can be defined as:
\begin{equation}
\begin{aligned}
   \mathcal{L}_{Prior} = -\ln p^{*} (\theta | \mathcal{D}) = \frac{1}{2} (\theta -\mu_{S}^{*})^{\top} \Sigma_{S}^{*{-1}} (\theta -\mu_{S}^{*}).
\end{aligned}
\label{BLoss}
\end{equation}

As mentioned in \cite{zhao2018federated}, separately solving subproblems on different clients leads to a large divergence in weights. This divergence will further lead to the problems of local forgetting and aggregation error as discussed in Sec. \ref{SEC:Introduction}.
As stated in \cite{lee2017overcoming,zenke2017continual}, the global covariances $\Sigma_{S}^{*{-1}}$ captures the correlations between parameters from the different local models and signifies their importance.
Although both the proposed prior loss and FedCurv \cite{shoham2019overcoming} have a similar formation, improving FedProx by treating the diagonal items of an empirical Fisher information matrix $\bar{F}$ as the anisotropic stiffness of parameters, they are totally different in their theories and implementations. 
Our prior loss stems from the PI strategy and aims to estimate an online-evaluated $\bar{F}$ while FedCurv merely tries to mitigate local forgetting by optimizing a decomposed global posterior $p(\theta | \mathcal{D} ) = p(\mathcal{D}_{n} | \theta) + p(\theta | \bar{\mathcal{D}}_{n} )$ during local training.
In practice, FedCurv is unable to reduce AE directly without using our proposed aggregation method. 
Furthermore, FedCurv will lead to a biased evaluation of $p(\theta | \mathcal{D} )$ since it approximates $p(\theta | \bar{\mathcal{D}}_{n} )$ using an offline matrix $\bar{F}$.

In addition, the loss $\mathcal{L}_{Task}$ corresponding to the tasks of local clients needs to be considered as well.
To sum up, the objective $\mathcal{J}_{n}(\theta)$ of client $n$ can be defined as the sum of the task loss $\mathcal{L}_{Task}$ and the proposed prior loss $\mathcal{L}_{Prior}$:
\begin{equation}
\begin{aligned}
   \mathcal{J}_{n}(\theta) = \mathcal{L}_{Task} + \lambda \mathcal{L}_{Prior}, \label{ALoss}
\end{aligned}
\end{equation}
where $\lambda$ is a balance factor that regulates the relative importance of the two losses.
Under the combined regularization of $\mathcal{L}_{Task}$ and $\mathcal{L}_{Prior}$, $ p(\theta|\mathcal{D}_{n})$ and $p^{*} (\theta | \mathcal{D})$ can be maximized at the same time, so that the corresponding local models perform well not only on that client but also on other clients.
Thus, we can see that the local forgetting issue has been effectively mitigated by our proposed method.

\subsection{Federated Online Laplace Approximation}\label{SEC:Estimation}

By applying the Bayesian framework from \cite{mackay1992practical}, we can treat parameters learned on client $n$ as the expectation of the posterior $p(\theta | \mathcal{D}_{n})$, and use the inverse of the average Hessian matrix $\bar{H}_{n}$ of the negative log posterior over data samples to approximate the covariance.
However, since the size of $\bar{H}_{n}$ is $d \times d$, where $d$ is the number of parameters, directly computing $\bar{H}_{n}$ has a time and storage complexity of $O(d \times d)$ in one client. 
Thus, we need to approximate $\bar{H}_{n}$ in a more efficient and practical way, as it is very expensive to calculate and store $\bar{H}_{n}$ directly.
Inspired by \cite{martens2010deep,lecun1990optimal}, we assume that the parameters in the covariance matrix are independent from each other. 
Therefore, we can approximate $\bar{H}_{n}$ using a diagonal matrix, thus decreasing the number of items in each covariance matrix from $O(d \times d)$ to $O(d)$. 

In addition, the diagonal approximation makes it is easy to calculate the inverse of $\Sigma_{n}$ or $\Sigma_{n}^{-1}$ in Eq. \ref{EQ:ModePro} because finding the inverse of a diagonal matrix is the same as taking the reciprocal of its diagonal elements.

Specifically, to calculate the covariance of $p(\theta | \mathcal{D}_{n})$, we utilize our own local training strategy PI to integrate both the likelihood function and the prior distribution $p(\theta)$. 
Using the approach from \cite{lee2017overcoming,zenke2017continual}, the likelihood function can be approximated by a second order regularization item $H_{n}$, which is the average Hessian matrix of $\mathcal{L}_{\text {Task}}=-\ln p\left(\mathcal{D}_{n} | \theta\right)$. 
Therefore, the posterior in Eq. \ref{BClientPI} can be further decomposed into:
\begin{eqnarray}
 \hspace{-1cm} \ln p(\theta | \mathcal{D}_{n})
                                &\approx& -\frac{1}{2} (\theta - \mu_{n})^{\top} \bar{H}_{n} (\theta - \mu_{n}) \\
                                &-& \frac{1}{2} (\theta - \mu_{S}^{*})^{\top} \Sigma_{S}^{*{-1}} (\theta - \mu_{S}^{*}) + C. \label{VPLP}
\end{eqnarray}

However, calculating the second derivative directly is also computationally expensive. 
Thus, we explore an empirical Fisher information matrix $\bar{F}$, which is a quadratic form for approximating the Hessian matrix. 
Such an approximation has been successfully used for natural gradient optimization \cite{pascanu2013revisiting}, which approximates the Hessian matrix for a mini-batch at each step to scale the gradient.
Similarly, for client $n$ with input $x$ and target $y$, we use the empirical Fisher matrices $\bar{F}_{n}$ of each training round to approximate $\bar{H}_{n}$ by:
\begin{equation}
  \bar{F}_{n} = \frac{1}{|\mathcal{D}_{n}|} \sum_{(x, y) \in \mathcal{D}_{n}} \nabla \log p(y \mid x,\theta) \nabla \log p(y \mid x,\theta)^{\top} .\label{AMu}
\end{equation} 
Eq.~\ref{AMu} is computable. In order to reduce the storage overhead to $O(d)$, a diagonal matrix of $\bar{\mathbf{H}}$ can be estimated by the diagonal of empirical $\mathbf{F}$. Although some information is lost by the absence of non-diagonal elements~\cite{mackay1992practical}, it is a common practice to use diagonal approximations for computational purposes ~\cite{lecun1990optimal,zenke2017continual,maddox2019simple, chen2020fedbe,kirkpatrick2017overcoming,ritter2018scalable}. We denote ${diag}(\cdot)$ as the diagonal of a matrix. 
\begin{equation}
\begin{aligned} \label{diagonal}
{diag}(\bar{\mathbf{H}}) & ={diag}(\mathbf{F}+\lambda \mathbf{I}) \\
& \approx \frac{1}{|\mathcal{D}|} \sum_{i=1}^{|\mathcal{D}|} {diag}\left(\nabla \log p(y \mid x,\theta) \nabla \log p(y \mid x,\theta)^{\top}\right)\\
&+\lambda \mathbf{I}
\end{aligned}
\end{equation}
Storing and computing Eq.~\ref{diagonal} for a neural network is efficient. It is equivalent to calculating the expectation of the square of the gradient of the loss function with respect to the parameters $\theta$. Meanwhile, ${diag}(\bar{\mathbf{H}})$ is easy to inverse, ${diag}(\bar{\mathbf{H}})^{-1}=\frac{1}{{diag}(\bar{\mathbf{H}})}$. Therefore, we select the diagonal of the empirical Fisher information matrix $diag(\bar{F}) $ to approximate $\bar{F}$:
\begin{equation}
\begin{aligned}
 diag(\bar{F}) = \frac{1}{|\mathcal{D}|} \sum_{(x, y) \in \mathcal{D}} sq(\nabla \log p(y \mid x,\theta)), \label{EqDiagF}
 \end{aligned}
\end{equation}
where $sq(\cdot)$ denotes a coordinate-wise square.

Nevertheless, as mentioned in \cite{martens2016second}, both $\bar{F}$ and $diag(\bar{F})$ are crude and biased approximations of $\bar{H}$, resulting in incorrect curvature of the posterior. 
For example, when certain parameters with small variances in the posterior are optimal, methods \cite{kirkpatrick2017overcoming,ritter2018scalable} applying the $diag(\bar{F})$ approximation around the optimal parameter will overestimate their variances. 
This is because the inverse of the variance obtained by Eq. \ref{EqDiagF} with $\nabla \log p(y \mid x,\theta) \approx 0$ around the optimal parameter is very small, leading to a large variance. 
Previous works \cite{zenke2017continual,schwarz2018progress} in the area of continuous learning have corrected this approximation error using an online approximation method. 
To this end, $ sq(\nabla \log p(y \mid x,\theta))$ is accumulated over the entire optimization process to approximate $\bar{H}$, instead of using an offline method to calculate $diag(\bar{F})$ around the optimal parameter $\theta^{*}$ after the optimization.
Similarly, we use $diag(\bar{F})_{n}$ averaged over all optimization steps from the initial step $t=1$ to the final step $t=T$ during local training on clients to approximate $\bar{H}_{n}$:
\begin{equation}
\begin{aligned}
 diag(\bar{F}) = \frac{1}{|\mathcal{D}| \cdot T} \sum_{t=1}^{T} \sum_{(x, y) \in \mathcal{D}} sq(\nabla \log p(y \mid x,\theta_{t})).  \label{EqAvgDiagF}
 \end{aligned}
\end{equation}

After taking the second derivative of $\ln p(\theta | \mathcal{D}_{n})$, an online-evaluated local covariance $\Sigma_{n}^{-1}$ can be obtained by the sum of  $\Sigma_{S}^{*{-1}}$ and $\bar{F}_{n}$:
\begin{equation}
   \Sigma_{n}^{-1} = - \mathbb{E} [\frac{\partial^{2} }{\partial \theta^{2}}\ln p(\theta | \mathcal{D}_{n})]  \approx \bar{F}_{n} + \Sigma_{S}^{*{-1}}. \label{EQ.UnbalancedLocalSigma}
\end{equation}
However, directly summing these will cause the covariance of the prior $p(\theta)$ of the next round to accumulate continuously, reducing the effect of the likelihood $p(\mathcal{D}_{n}|\theta)$, which is also very important for local learning. 
In order to balance their effects, we scale the covariances using the round index $r$ as:
\begin{equation}
   \Sigma_{n}^{-1} \approx \frac{1}{r} \bar{F}_{n} + \frac{r-1}{r} \Sigma_{S}^{*{-1}}.  \label{EQ.BalancedLocalSigma}
\end{equation}

Furthermore, by applying Eq. \ref{EQ.BalancedLocalSigma} during the aggregation stage at round $r=R$, we can obtain an online evaluated global covariance $\Sigma_{S,R}$:
\begin{eqnarray}
\Sigma^{-1}_{S,R} = \sum_{n} \pi_{n} \Sigma^{-1}_{n,R} = \frac{1}{R} \sum_{r=1}^{R} \sum_{n} \pi_{n} \bar{F}_{n,r} +  \gamma I,
\label{EQ.OnlineGlobalSigma}
\end{eqnarray}
where the initial prior $p_{r=1}(\theta) \approx \mathcal N(O, \gamma I)$ is given with zero expectation and an identity covariance scaled by a constant $\gamma$. 
Eq. \ref{EQ.OnlineGlobalSigma} can also be derived directly utilizing Laplace approximations and PI when approximating the global posteriors, as shown in the appendix. 

Our federated online Laplace approximation provides an effective iteration method for estimating the probabilistic parameters of both the global and local models in an online manner.
The probabilistic parameters obtained by our method can be directly used in the FL framework.
In contrast, because the probabilistic parameters are always overestimated by offline methods, as discussed in Eq. \ref{EqDiagF}, these methods are generally ineffective for the FL task in practice.


\subsection{Algorithm  Complexity and Privacy Security Analysis}

\textbf{Algorithm  complexity.}
Although our method requires more computational steps, it has the same algorithm complexity as the standard FedAvg.
On the client side, the back-propagation algorithm is used to train the model in FedAvg, whose computational and space complexity are both $O(d)$ as we know.
During local training, compared with FedAvg, our framework only needs to additionally accumulate the gradient of each parameter to calculate the local covariance. As discussed in Sec. \ref{SEC:Estimation}, the complexity of this extra operation is $O(d)$.
Therefore, the overall algorithm complexity of our framework running on the client is $O(d)$.

On the server side, FedAvg averages the parameters of $n$ neural networks to aggregate local models, which has a complexity of $O(n \times d)$.
In the aggregation step, our framework first averages the inverses of the local covariance of $n$ clients, and then takes the inverse of this to obtain a global covariance.
Benefiting from our FOLA algorithm described in Sec. \ref{SEC:Estimation}, each inverse of the local covariance is a diagonal matrix, so the averaging and inverse operations require $O(n \times d + d)$ complexity.
With the global covariance, our aggregation method sums the matrix multiplications between each inverse of the diagonal local covariance and the corresponding local expectation, and then applies matrix multiplication between the global covariance and this sum.
These operations require a complexity of $O(2 \times n \times d + d)$.
Therefore, our aggregation method has an algorithm complexity of $O(3 \times n \times d + 2 \times d)$ which is equivalent to $O(d)$ because $n$ is much smaller than $d$.

In summary, the algorithm complexity of our framework is $O(d)$, which is the same complexity as FedAvg.


\textbf{Privacy security.}
Similar to the classical FedAvg algorithm, our method protects the privacy of each component in the system by sharing only gradient-related information between the central server and clients.
To update the parameters, an additional Fisher information matrix, estimated by the square of the gradient, is merely transferred.
It is worth noting that the gradient information cannot be recovered from the Fisher information matrix, as obviously we cannot obtain the gradient from its square.
Thus, recently proposed attack techniques like \cite{zhu2019deep} cannot leak sensitive data or jeopardize the privacy of our system.
Therefore, in terms of privacy, our method is not significantly different from the classical FedAvg algorithm.
Moreover, to further enhance the protection of data privacy, we can easily to apply secure aggregation methods \cite{bonawitz2016practical} to our method, as successfully done for FedAvg.

\begin{algorithm}[]
\SetAlgoLined
\SetAlgoNoLine
\caption{A Bayesian Federated Learning Framework with Multivariate Gaussian Product}

$\theta$: model parameters \\
$\mu$: mean of prior distribution of $\theta$ \\
$\Sigma$: covariance of prior distribution of $\theta$ \\
$\lambda$: factor of regular penalty term \\
\BlankLine
\textbf{Server executes:} \\
Initialize $\theta$, $\mu_{S}$ and $\Sigma_{S}$ \\
\For{each round r = 1,2,...} {
    \For{each client $n$ from 1 to $N$ in parallel} {
        $\mu_{n}$, $\Sigma_{n}$ = ClientUpdate($\mu_{S}$, $\Sigma_{S}$, r) \\
    }
    $\Sigma_{S}^{-1} = \sum_{n}^{N} \pi_{n} \Sigma_{n}^{-1} $ \\
    $\theta = \mu_{S} = \Sigma_{S} \sum_{n}^{N} \pi_{n} \Sigma_{n}^{-1} \mu_{n}$ \\
}
\BlankLine
\textbf{ClientUpdate}\ ($\mu_{S}$, $\Sigma_{S}$, r): \\
$\theta= \mu_{S}$ \\
Initialize $\mu_{n}$ and $\Sigma_{n}$ \\
\For{each local epoch $i$ from 1 to $E$} {
    $B$ = split $\mathcal{D}_{n}$ into batches of size $B$ \\
    \For{$b$ in $B$} {
        $g = \frac{\partial \mathcal{L}_{task}}{\partial \theta}$ \\
        $diag(\Sigma_{n}) = diag(\Sigma_{n}) + diag(gg^{\top})$ \\
        $\mathcal{L}_{prior} = \frac{1}{2} (\theta-\mu)diag(\Sigma_{S}) (\theta-\mu)^{\top}$ \\
        $\theta \leftarrow \theta - lr \cdot (g + \lambda \frac{\partial \mathcal{L}_{prior}}{\partial \theta})$ \\
    }
}
$\mu_{n}=\theta$ \\
$\Sigma_{n}= \frac{1}{r}\Sigma_{n} + \frac{r-1}{r}\Sigma_{S}$ \\
\Return $\Sigma_{n}$, $\mu_{n}$ to server

\label{alg:Framwork}
\end{algorithm}

\section{Experiments\label{SEC:EX}}

We compare our proposed framework with several baselines \cite{mcmahan2016communication,shoham2019overcoming} on the MNIST \cite{lecun1998gradient} and CIFAR-10 \cite{krizhevsky2009learning} datasets.
We train a multilayer perceptron (MLP) network on MNIST and a convolutional neural network (CNN) on CIFAR-10.
In order to study the AE issue, we evaluate the global accuracy (GA) of the global model for all experiments.
The better the performance of the global model, the smaller the AE of the aggregation process. 
Besides, we report the average local accuracy (LA) of local models on  CIFAR-10 to analyze LF.
LA is defined as $LA=\sum_{n} \pi_{n}  Precision(\theta_{n})$, where $Precision(\theta_{n})$ is the precision evaluated on a global test set using corresponding local parameters $\theta_{n}$, and $\pi_{n}$ is the proportion of the size of the local training set to the whole training set.
Generally, algorithms with larger LA suffer from less LF.
We compare the empirical results of our framework with other popular FL methods in terms of the degree of heterogeneity and number of communication rounds.


\subsection{Experimental Setup} 

In order to measure the effects of different heterogeneous degrees of data, we use a Dirichlet distribution controlled by a concentration parameter $\alpha > 0$ to generate populations of data with different heterogeneities. 
This setting is the same as in \cite{wang2020federated}.
In a population of $N$ clients, the class labels follow a categorical distribution over $K$ classes parameterized by a vector $\bm{q}_{n}$ ($\bm{q}_{n,i} \geq  0, n \in [1,N] , i \in [1, K] \ \text{and} \  ||\bm{q}||_{1} = 1$). 
For client $n$, $\bm{q}_{n}$ is sampled from a Dirichlet distribution $\bm{q}_{n}$ $\sim$ Dir($\alpha \cdot \bm{p}$), where \bm{$p$} characterizes a uniform class distribution over $K$ classes.

In addition, because the client activation rate affects the evaluation of AE, we set it to 1 for all experiments. That is, all clients will participate in each round of training period. 
SGD is chosen as the optimizer for all experiments.

\begin{table}[t]
\centering
\caption{Data augmentation of CIFAR10-CNN model. For AutoAugment and Cutout, we use the codes provided in \cite{cubuk2019autoaugment} and \cite{devries2017improved}, respectively.}
\begin{tabular}{c}
\hline
\textbf{Data Augmentation Processes}        \\ \hline
RandomCrop (32, padding=4, fill=128)         \\
RandomHorizontalFlip ()                      \\
AutoAugment for CIFAR10                     \\
Cutout (n\_holes=1, length=16)               \\
Normalize ((0.5, 0.5, 0.5), (0.5, 0.5, 0.5)) \\ \hline
\end{tabular}
\label{TABDA}
\end{table}

\subsubsection{MNIST-MLP model}
For the experiment on the MNIST, we employ a 784-500-300-10 MLP with different hyperparameters: local training epochs $E \in \{1,5,10,15,20\}$, learning rates $lr\in \{1, 10^{-1}, 10^{-2}, 10^{-3}, 10^{-4}\}$, batch sizes $B\in \{8, 32, 64, 128, 254\}$, data heterogeneity degrees $\alpha \in \{0.01, 0.1, 1, 10, 100\}$  and client numbers $N \in \{5,10,20,50,100\}$. 
This model achieves 97.57\% accuracy when trained in a centralized manner.

In this experiment, we compare the results of different aggregation methods in the first round of training over various hyperparameters.
Before the first round of training, all local parameters are identical in terms of coordinates, but will be optimized during the following training.
Since the distributions of local data are generally different, the positions of modes of different local models have great divergence after local training, resulting in aggregation error. 
Therefore, in the first round, we are able to study the problem of aggregation error with respect to different settings and evaluate the abilities of different algorithms to reduce it.

\begin{figure}[t]
\centering
\includegraphics[width=7cm]{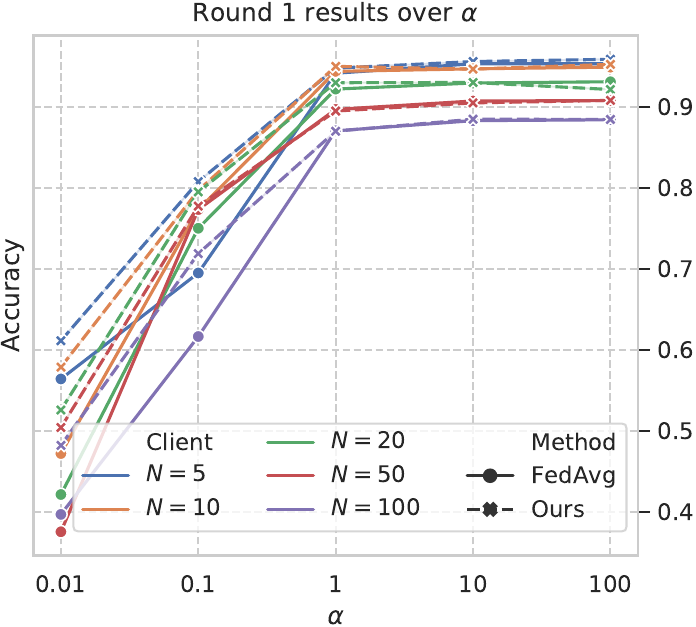}
\caption{The best accuracy in each case of $\alpha$ and $N$. Different colors indicate different numbers of clients. The Soild line is FedAvg and dashed line is our model.}
\label{mlp_a}
\end{figure}

\begin{figure}[t]
    \centering
    \subfigure[$N=10$]{
	\includegraphics[width=1.55in]{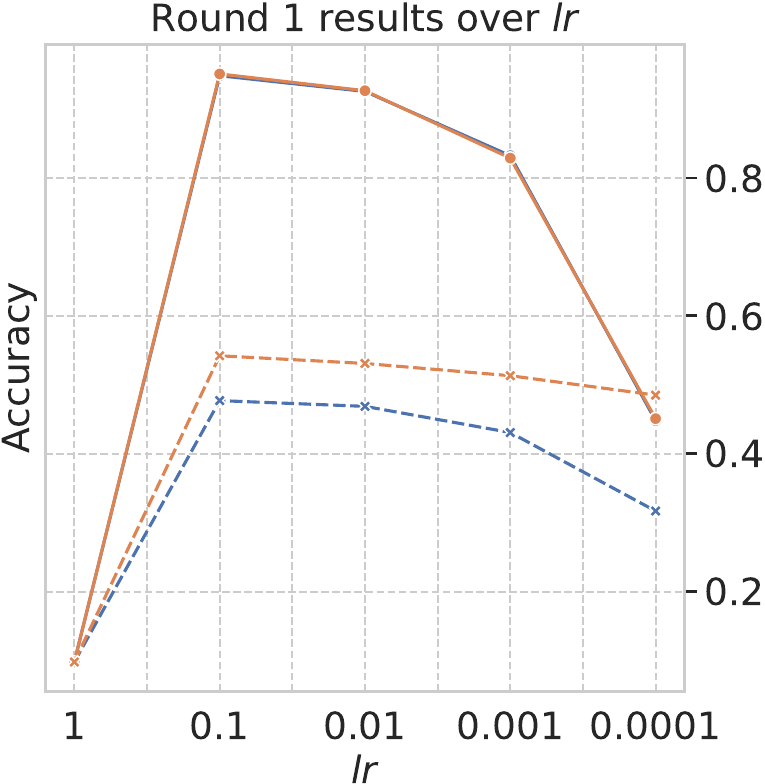}
        \label{mlp_b}
    }
    \centering
    \subfigure[$N=10$]{
    	\includegraphics[width=1.55in]{ 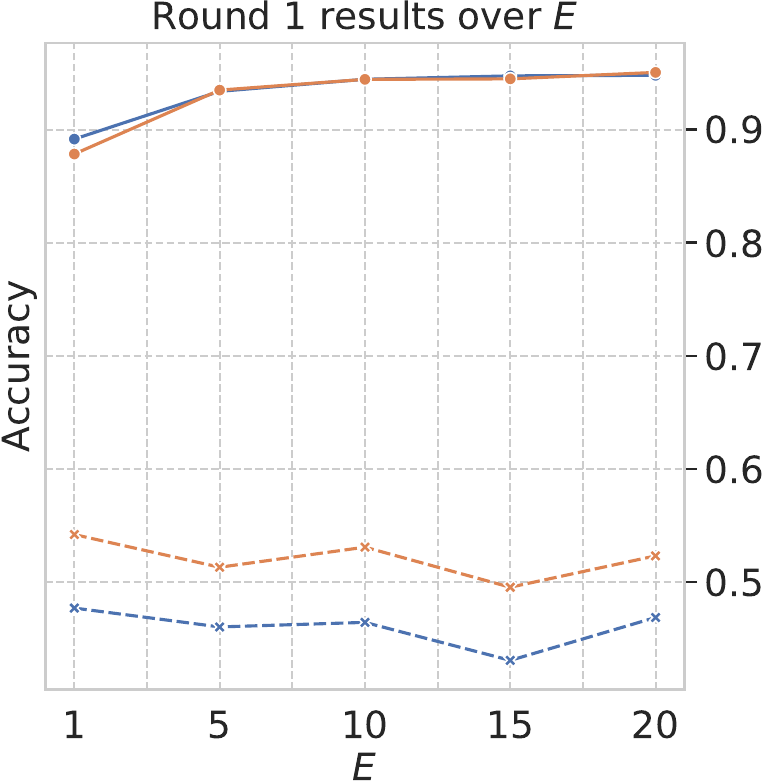}
        \label{mlp_c}
    }
    \centering
    \subfigure[$N=10, E=1, lr=0.1$]{
	\includegraphics[width=3.3in]{ 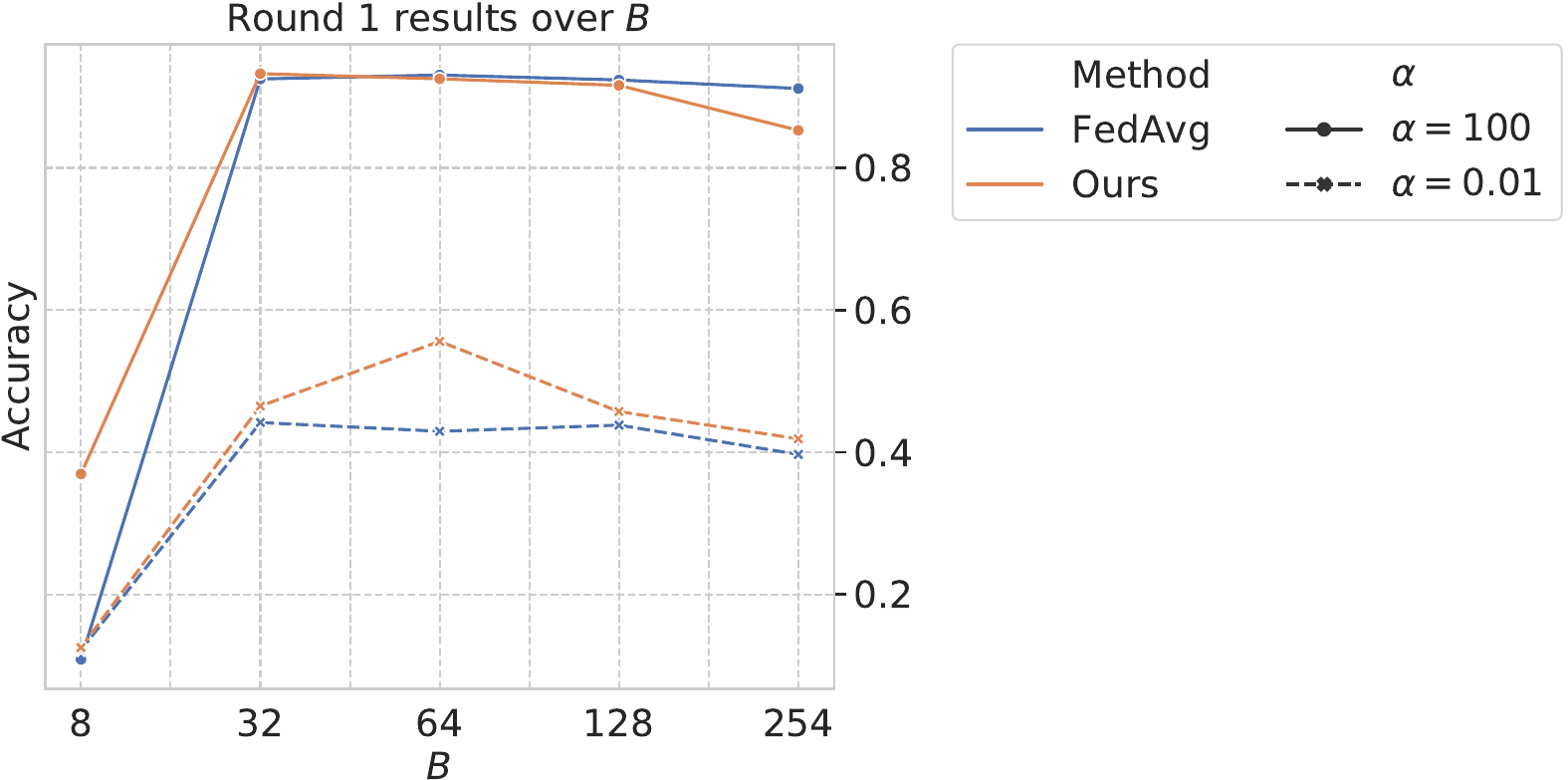}
        \label{mlp_d}
    }
    \caption{Variation curves of GA for the MNIST-MLP model under different hyperparameters. 
    {Under the fixed hyperparameters in the captions, averaged results are plotted for other best hyerparameters.}
    Different colors indicate different methods, and different lines represent different values of $\alpha$.}
    \label{fig.mlp_hp}
\end{figure}

\begin{figure}[t]
\centering
\includegraphics[width=8cm]{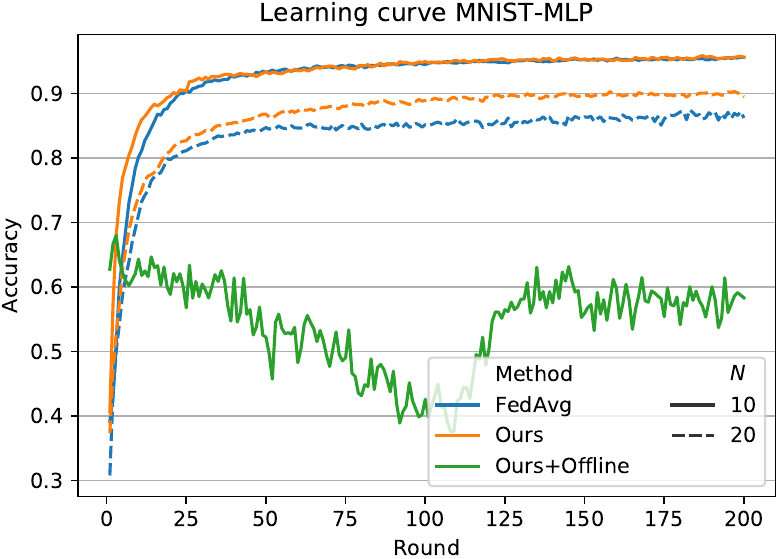}
\caption{Learning curves of the MNIST-MLP model with different client numbers $N$ under $\alpha=0.01$, $E=5$ and $lr=0.01$.
Different colors indicate different methods, and different lines represent different numbers of clients.}
\label{FIGMLPC}
\end{figure}

We show the best results for the first round of training in Fig. \ref{fig.mlp_hp}. 
A grid search method is used to find the best learning hyperparameters $lr, E, B$ under various environment settings $(N, \alpha)$. 
Each group of experiments is run 30 times, and the average results are taken to avoid the influence of randomness.
Additionally, we draw the learning curves of the proposed method using the MNIST-MLP model in Fig. \ref{FIGMLPC} to demonstrate the effect of AE on the learning process. 
We run all experiments on the MNIST-MLP model with two GPUs (2 x Nvidia Tesla v100 16GB), requiring nearly six days.

\subsubsection{CIFAR10-CNN model} 

We use the same architecture as \cite{mcmahan2016communication} in the CIFAR10-CNN model for the experiments on CIFAR10.
When trained in a centralized manner, this CIFAR10-CNN model achieves 86.24\% accuracy, which is not the state-of-the-art on CIFAR10, but it is sufficient to show the relative performances for the purposes of our investigation.
The parameter $\lambda$ for FedCurv \cite{shoham2019overcoming} is searched from $\{1, 10, 10^{2}, 10^{3}, 10^{4} \}$.
In addition, all methods are run on clients with the following settings: batch size $B = 32$, number of clients $N=20$, local epoch counts $E \in \{1, 10, 20, 40, 80, 160\}$, learning rates $lr\in \{1, 10^{-1}, 10^{-2}, 10^{-3}, 10^{-4}\}$ and heterogeneity degrees of data $\alpha \in \{0.01, 0.1, 1, 10, 100\}$ for a total of 100 communication rounds.
As usual, some augmentation techniques are applied to the CIFAR10 dataset, the details of which are given in Tab. \ref{TABDA}.

\begin{figure*}[t]
    \centering
    \subfigure[GA over E under $\alpha=0.01$]{
        \includegraphics[width=2.2in]{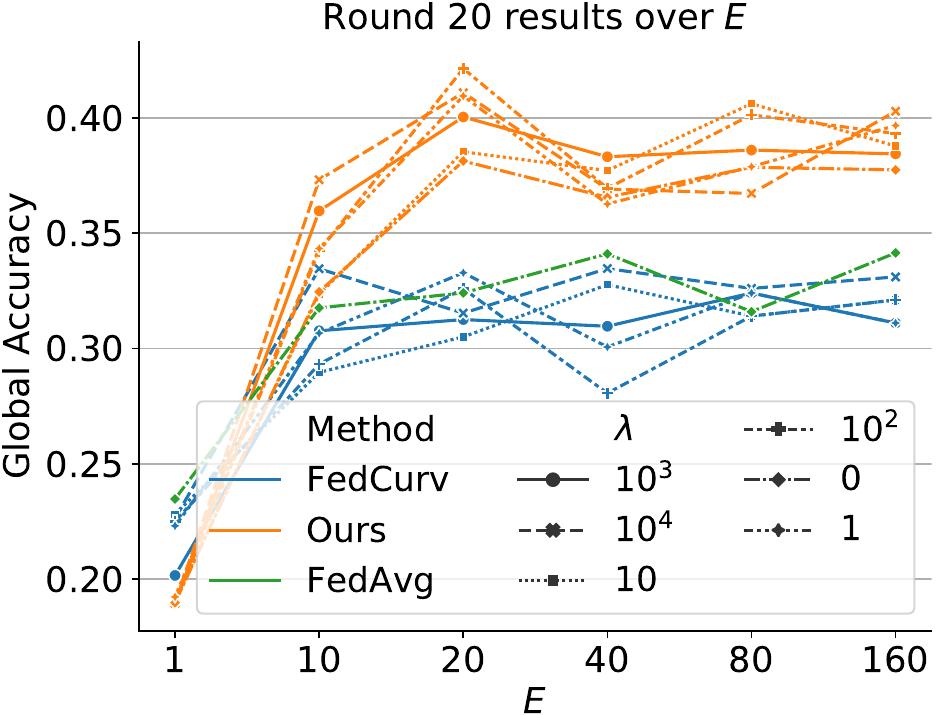}
        \label{cnn_GA_E}
    }
    \subfigure[LA over E under $\alpha=0.01$]{
	\includegraphics[width=2.3in]{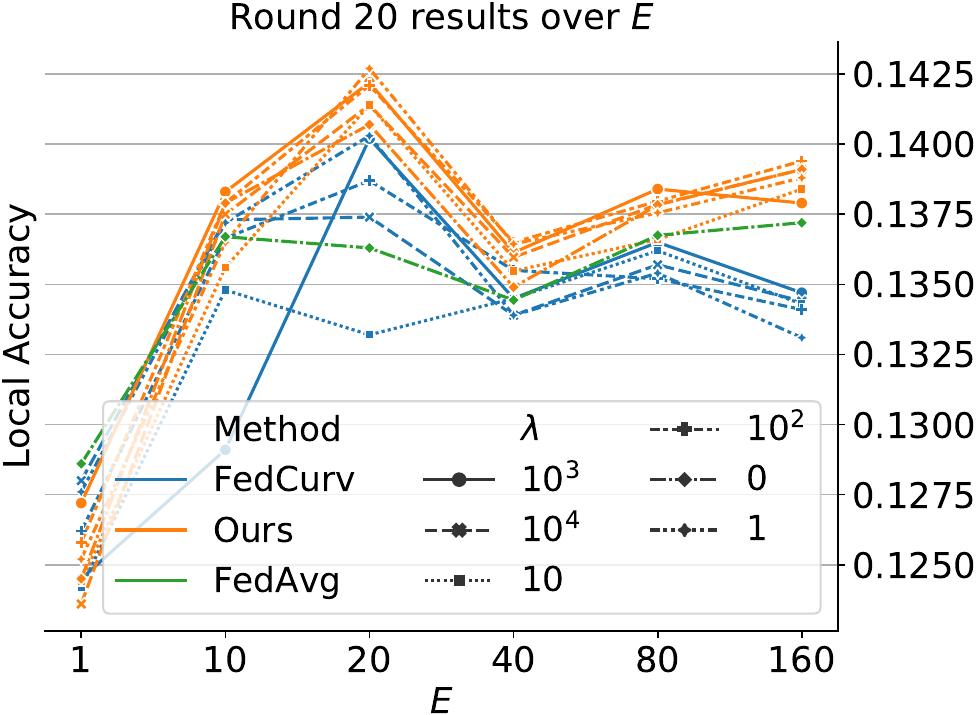}
        \label{cnn_LA_E}
    }
    \subfigure[Best accuracy in each case]{
    \includegraphics[width=2in]{ 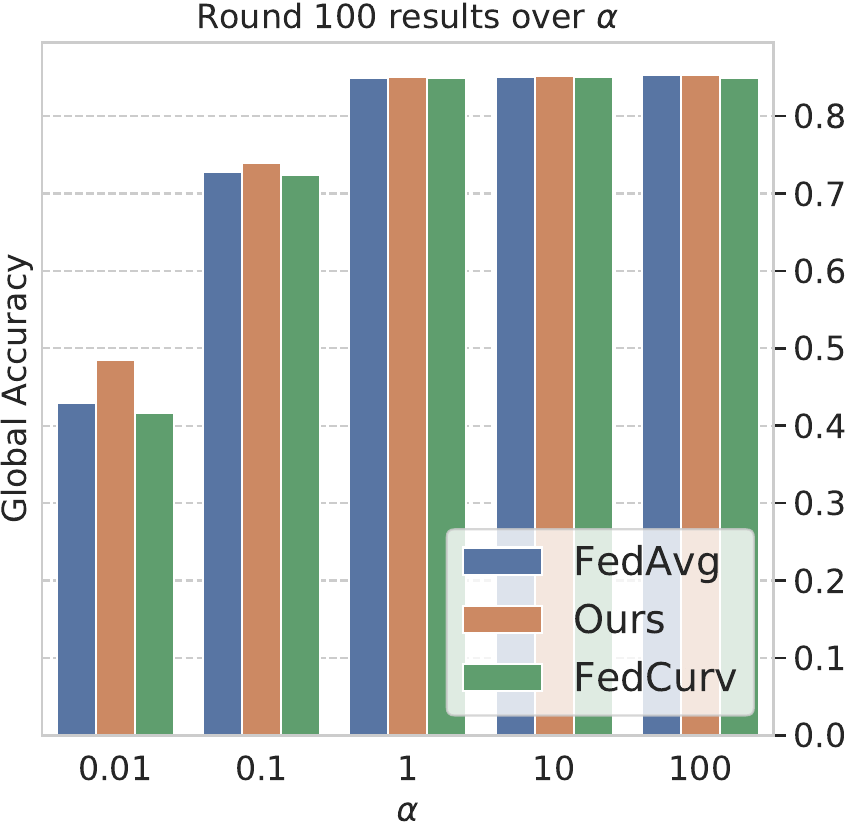}
        \label{cnn_a}
    }
    \caption{ GA and LA illustration under different settings of $E$ and $\lambda$ after running 20 rounds with $\alpha=0.01$. Different colors indicate different methods, and different lines represent different values of $\lambda$.
    (a) Global accuracy. 
    (b) Local accuracy.
    (c) We use the best hyperparameters searched from the first twenty rounds for further training until 100 rounds, and then calculate the GA.}
    \label{fig.cnn_GA_LA_E}
\end{figure*}

\begin{figure}[t]
\centering
\includegraphics[width=8cm]{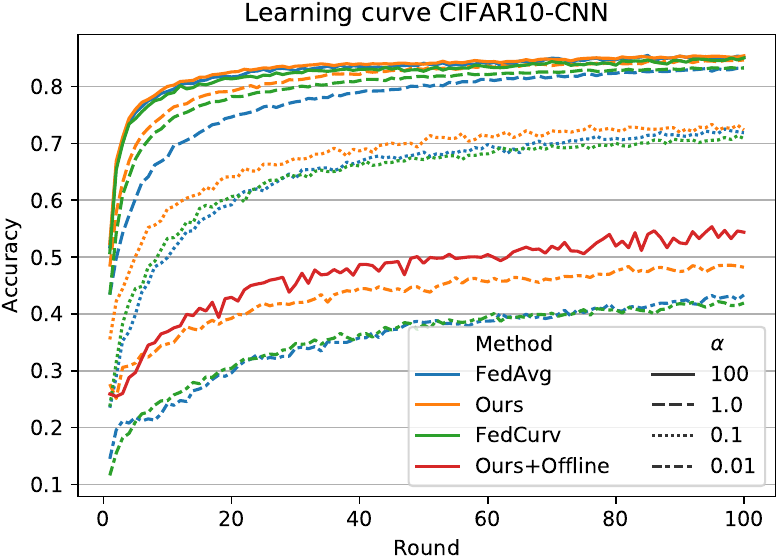}
\caption{Learning curves of the CIFAR10-CNN model with different degrees of data heterogeneity $\alpha$ for the case of $N=20$. 
The best $lr$ for all cases is $0.01$.
Different colors indicate different methods, and different lines represent different values of $\alpha$.}
\label{fig.cnn}
\end{figure}

We show the learning curves and best results in Fig. \ref{fig.cnn} and \ref{cnn_a} with different FL frameworks, including the proposed method, FedAvg and FedCurv. 
The grid search method is applied to find the best learning hyperparameters $lr$ and $E$ under various environment settings $\alpha$. 
We run all experiments of the CIFAR10-CNN model on an Nvidia DGX-2 platform and use eight GPUs (8 x Nvidia Tesla v100 32GB), requiring 15 days in total.




\subsection{Study on Aggregation Error}



According to their types of influences, we group factors affecting AE into two categories: environment hyperparameters and learning hyperparameters.
The environment hyperparameters $\alpha$ and $N$ establish the physical environment of FL.
For a given model architecture and dataset, the true local posterior distribution can be determined only when the two parameters are fixed.
Thus, they directly affect the divergence of the local posteriors.
However, the learning hyperparameters $B$, $E$, and $lr$ also impact the local training process in a given FL environment. 
They indirectly affect AE by controlling the step size and the direction of the parameter optimization during local training.
By analyzing the changes in AE under these two groups of factors, we can achieve a deeper understanding of how AE occurs and affects the final accuracy.

\begin{figure}[t]
\centering
\includegraphics[width=8cm]{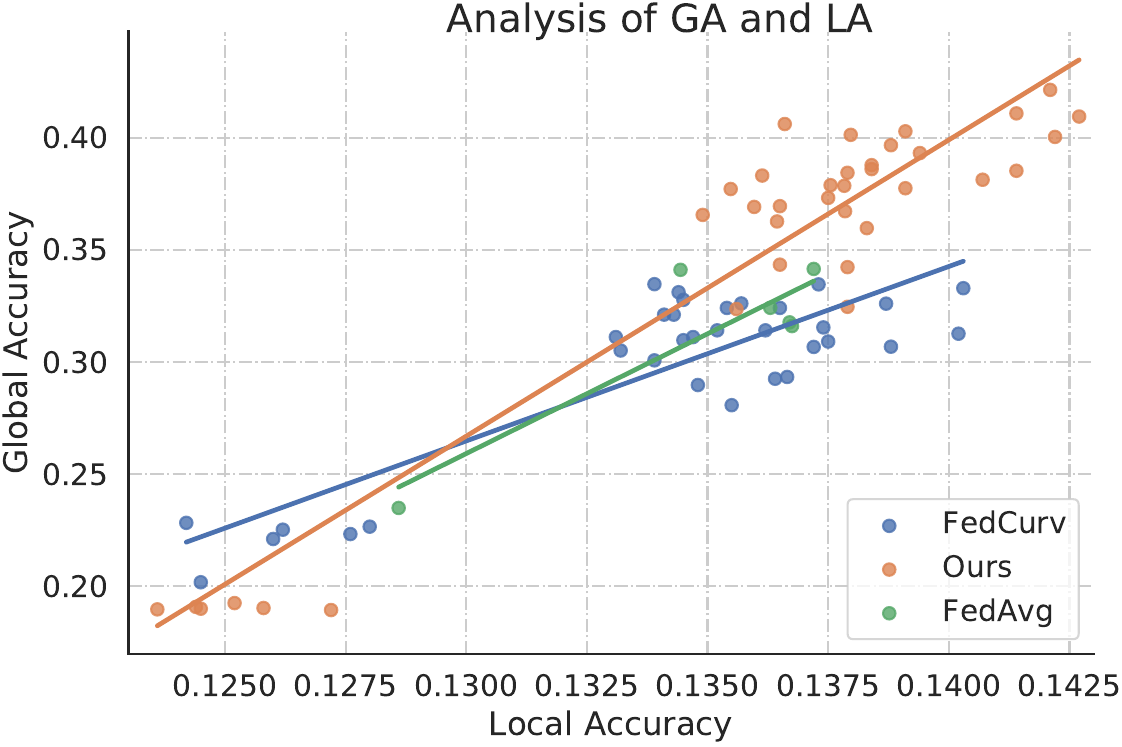}
\caption{Correlation analysis of all GA and LA results for the hyperparameter search under $\alpha=0.01$. Different colors indicate different methods. }
\label{Regression}
\end{figure}

\textbf{Effect of environment hyperparameters.}
As observed from Fig. \ref{mlp_a}, AE has a strong relationship with the parameters $\alpha$ and $N$.
First, when $\alpha$ decreases, AE increases accordingly.
In other words, AE is negatively correlated to $\alpha$.
In contrast, when $N$ becomes larger, AE will always increase accordingly. 
These correlations becomes more obvious when $\alpha$ is between 1 and 100. 
Besides, AE is different for various methods.
When $\alpha$ is small, FedAvg suffers from a large AE. 
This is because, as discussed in Sec. \ref{SEC:Method}, a smaller $\alpha$ makes the local posteriors more heterogeneous so it is difficult for FedAvg to collapse their mixture.
However, in contrast, AE can be significantly reduced by our method, even when $\alpha$ is extremely small.
This experimental phenomenon is consistent with our analysis in Sec. \ref{SEC:Introduction} and \ref{SEC:Method}, illustrating the superiority of our method.

\textbf{Effect of learning hyperparameters.} 
As shown in Fig. \ref{mlp_b}, \ref{mlp_c} and \ref{mlp_d}, the parameters $B$, $E$ and $lr$ are critical for the aggregation results but they are not absolutely related to AE.
Only when they are moderate can all the FL methods achieve low AE.
If $lr$ is too large or $B$ is extremely small, the local optima will become more divergent because both the step size and the direction of the optimization process vary.
As a result, the local parameters become heterogeneous after the local training, resulting in a large AE and a small GA.
Additionally, compared with FedAvg, our method can significantly improve the best GA curves as shown in Fig. \ref{mlp_b} and \ref{mlp_c} when the data is heterogeneous.
The performance improvement comes from our aggregation strategy, which always achieves a higher global posterior probability than FedAvg, as discussed in Fig. \ref{SEC:Method}.

\subsection{Comparison and Analysis}

\begin{table}[]
\centering
\caption{Results of CIFAR10-CNN under $\alpha=0.01$. The last two columns are the round numbers when the aggregated model achieves the corresponding global accuracies.}
\begin{tabular}{clclclclclclclc}
\hline
\multirow{2}{*}{Method} &
\multirow{2}{*}{$E$} & \multicolumn{2}{l}{Accuracy} & \multicolumn{2}{l}{Round (GA=)} \\ \cline{3-6} 
        &        & $GA$ & $LA$ & $30\%$ & $40\%$ \\ \hline
Ours    &     $20$   & $ 48.57 \%$   &  $ 14.57 \%$  & 4   & 21   \\
FedAvg  &    $40$  &  $ 42.16 \%$  &  $ 14.15 \%$  & 23   & 73   \\
FedCurv &   $10$    &  $ 41.31 \%$  &  $ 14.51 \%$  & 21   &  74  \\ 
FedNova &   10 & 10.00\% & 10.00\% & - & - \\
SCAFFOLD&   1 & 20.30\% & 13.71\% & - & - \\
FedBE &   20 & 40.02\% & 14.04\% & 42 & 90 \\
\hline
\label{TAB.CIFAR10}
\end{tabular}
\end{table}

Fig. \ref{mlp_a} and \ref{cnn_a} clearly show that our method achieves higher accuracy than other FL methods under various degrees of data heterogeneity.
Overall, all methods performs very similarly for large $\alpha$.
In contrast, with the decay of $\alpha$, the global accuracy of all methods falls, however our method only decreases slightly.
As shown in Fig. \ref{fig.cnn} and \ref{cnn_a}, when $\alpha=0.01$ in the CIFAR10-CNN model, our method achieves $48.57\%$ global accuracy, which is $5\%$ higher than others. 
In fact, the best GA scores of FedAvg and FedCurv are only 43.34\% and 42.16\%, respectively. 
In addition, from Fig. \ref{FIGMLPC} and Tab. \ref{TAB.CIFAR10}, we can see that our method also achieves a faster convergence rate than FedAvg. 
We observed that despite hyperparameter tuning, such as local epoch counts $E$, both FedNova and SCAFFOLD performed poorly under such extremely imbalanced conditions, which aligns with observations from other research studies~\cite{li2022federated}. The performance of FedBE was also suboptimal, potentially due to additional LF induced by server-side training, leading to a decline in its LA compared to FedAvg.

\textbf{Effect of mitigating local forgeting.}
As discussed in Sec. \ref{SEC:Introduction} and \ref{SEC:PriorLoss}, mitigating LF indirectly can reduce AE as well.
From Fig. \ref{Regression}, we can obtain similar observations.
Specifically, Fig. \ref{Regression} demonstrates that both LA and GA are generally positively-correlated.
In this experiment, the environment parameters $(N, \alpha)$ are fixed, and only the learning parameter is changed over the data points.
Therefore, LF can be improved with suitable learning parameters, making GA become larger in the next round of aggregation.
The phenomenon in Fig. \ref{Regression} sugessts that GA can be also reduced by minor LF.
In summary, it is practical to improve the performance of FL by mitigating LF with regularization methods.

\textbf{Learning hyperparameter sensitivity.}
In this experiment, we employ our aggregation strategy, but without the use of the improved local training method. 
The effect of learning hyperparameters $E, lr, B$ on our method is similar to their effect on FedAvg, as shown in Fig. \ref{fig.mlp_hp}.
However, the local CIFAR10-CNN models require longer training times for optimization than the MNIST-MLP model. 
As shown in Fig. \ref{fig.cnn}, in this case, we also observe that our method always achieves better results when $E$ increases.
Additionally, we find that FedCurv only performs better than FedAvg when $\alpha$ is a moderate value. 
When setting $\alpha=1$, FedCurv with $\mu=10^4$ converges faster than FedAvg. 
If $\alpha$ is extremely small, FedCurv suffers from a lower convergence rate than FedAvg.
As shown in  Fig. \ref{fig.cnn}, the dramatic fluctuations of FedCurv over $\alpha$ and $\mu$ indicate the unstable effect of indirectly reducing AE during local training.
In contrast, our method can directly reduce AE and then improve GA under various settings.

\begin{figure*}[t]
    \centering
        \subfigure[$\alpha=100$]{
        \includegraphics[width=2.1in]{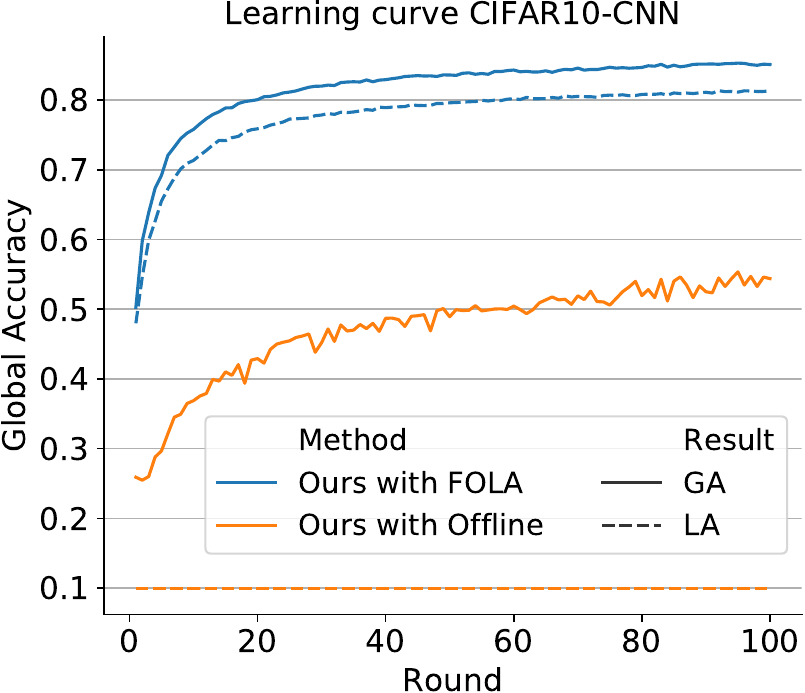}
        \label{cnn_GA_offline}
    }
    \subfigure[$\alpha=0.01$]{
        \includegraphics[width=2.15in]{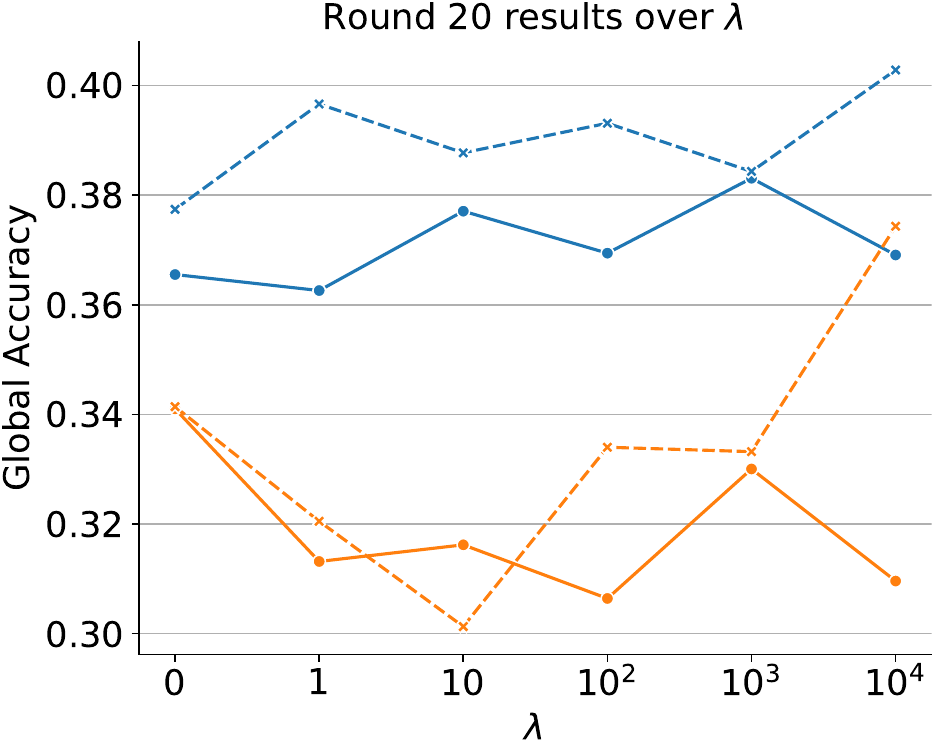}
        \label{cnn_GA_lambda}
    }
    \subfigure[$\alpha=0.01$]{
	\includegraphics[width=2.35in]{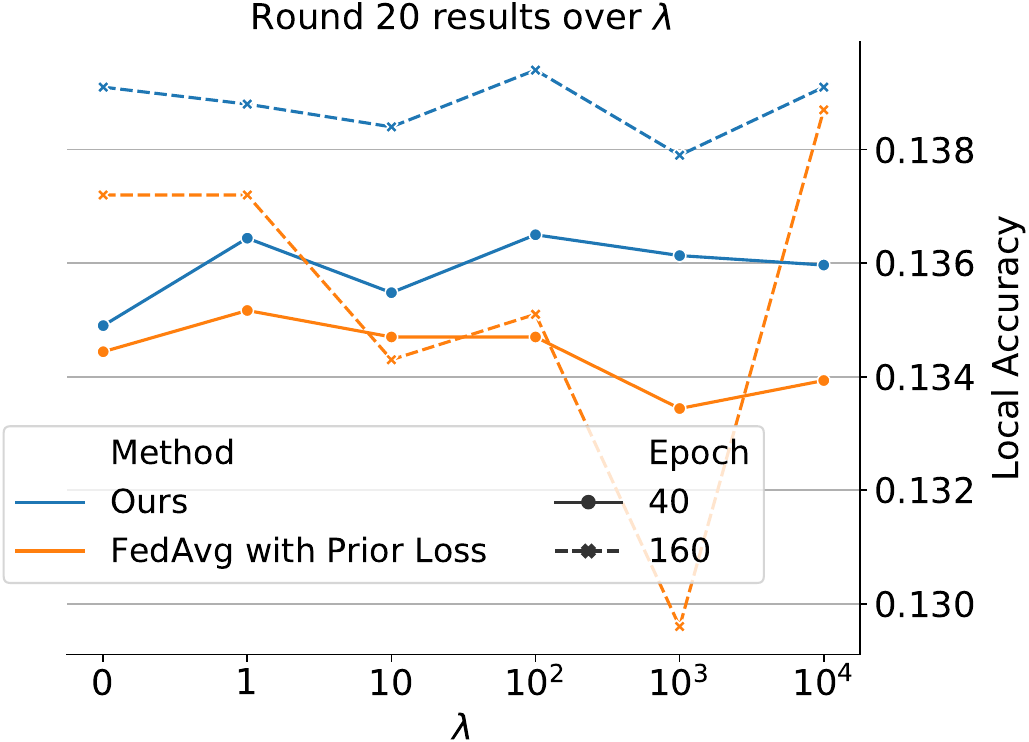}
        \label{cnn_LA_lambda}
    }
    \caption{(a) A comparison between FOLA and the offline version of our method without the prior loss. (b) Global accuracy comparison between our method and FedAvg with the prior loss. (c) Local accuracy comparison between our method and FedAvg with the prior loss.}
    \label{fig.cnn_GA_LA_lambda}
\end{figure*}

\subsection{Ablation Study}

In this section, we conduct ablation experiments on our framework to study the effect of our aggregation strategy, local strategy and online Laplace approximations. 

Firstly, based on our aggregation strategy, we compare two different Laplace approximations: FOLA and the offline method.
In Fig. \ref{FIGMLPC} and \ref{cnn_GA_offline}, we can see that the learning curves of the offline method vary dramatically after a few rounds. 
The inherent reason lies in the fact that the variance obtained by the offline method is crude and biased. 
Hence, the error in the evaluation of variance causes the aggregation parameters to deviate from the true optimal value.
In contrast, the variance obtained by our method is closer to the true variance and makes the learning process converge smoothly.
This experiment illustrates that the posterior probabilistic parameters obtained by FOLA are effective in FL, while those obtained by the offline method employed in \cite{kirkpatrick2017overcoming,shoham2019overcoming} are difficult to use. 
Additionally, we provide a deeper study of different Laplace approximations in Sec. \ref{EX.LAP}. 

Next, we keep our local training strategy unchanged and compare two aggregation strategies, including the method used in FedAvg and our algorithm.
The covariance is evaluated by FOLA. 
As shown in Fig. \ref{cnn_GA_lambda} and \ref{cnn_LA_lambda}, our aggregation strategy boosts the accuracy curve while FedAvg achieves lower accuracy than ours.
As well known, parameters with a higher posterior achieve better test accuracy, implying that our aggregation method can obtain model parameters with higher posterior than FedAvg.
Additionally, we observe that using our full framework can achieve better performance than only using a single module.
Therefore, we conclude that simultaneously using our aggregation and local training strategies can further improve the performance.

\subsection{Effect of Laplace Approximations on Aggregation} \label{EX.LAP}

\begin{figure*}[t]
    \centering
        \subfigure[The covariance in a normalized $\Sigma_{1}^{-1}$.]{
        \includegraphics[width=2.3in]{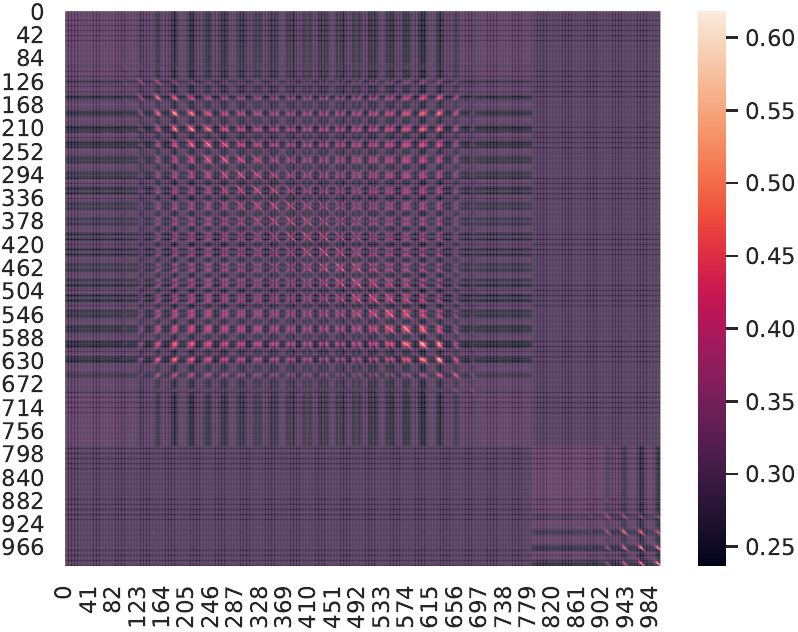}
        \label{FullHessian}
    }
    \subfigure[]{
        \includegraphics[width=2.2in]{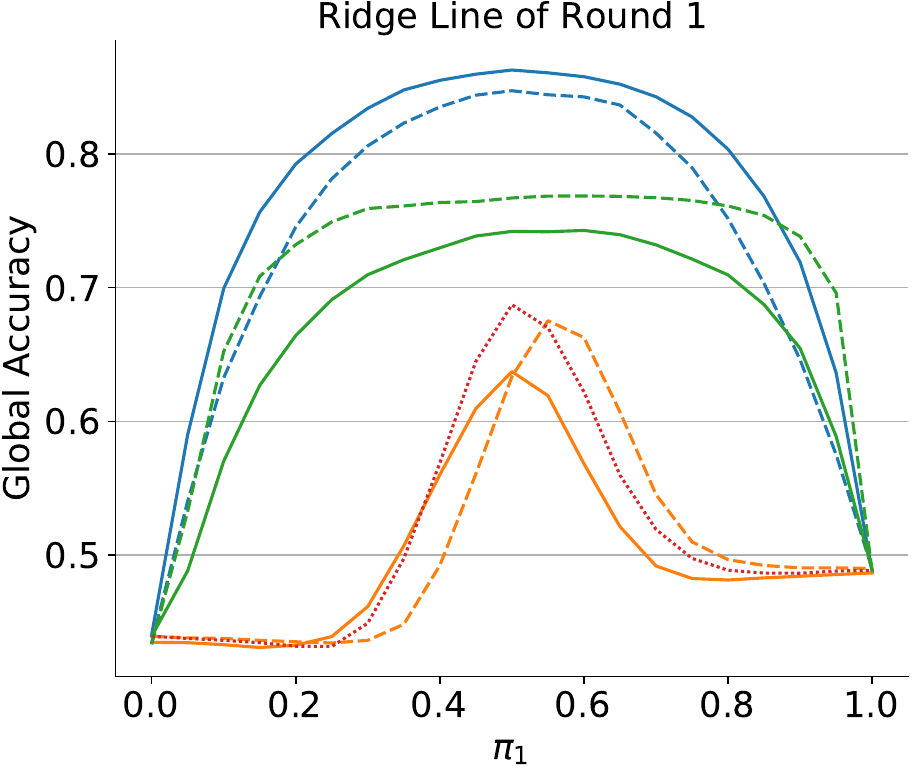}
        \label{Ridge1}
    }
    \subfigure[]{
	\includegraphics[width=2.2in]{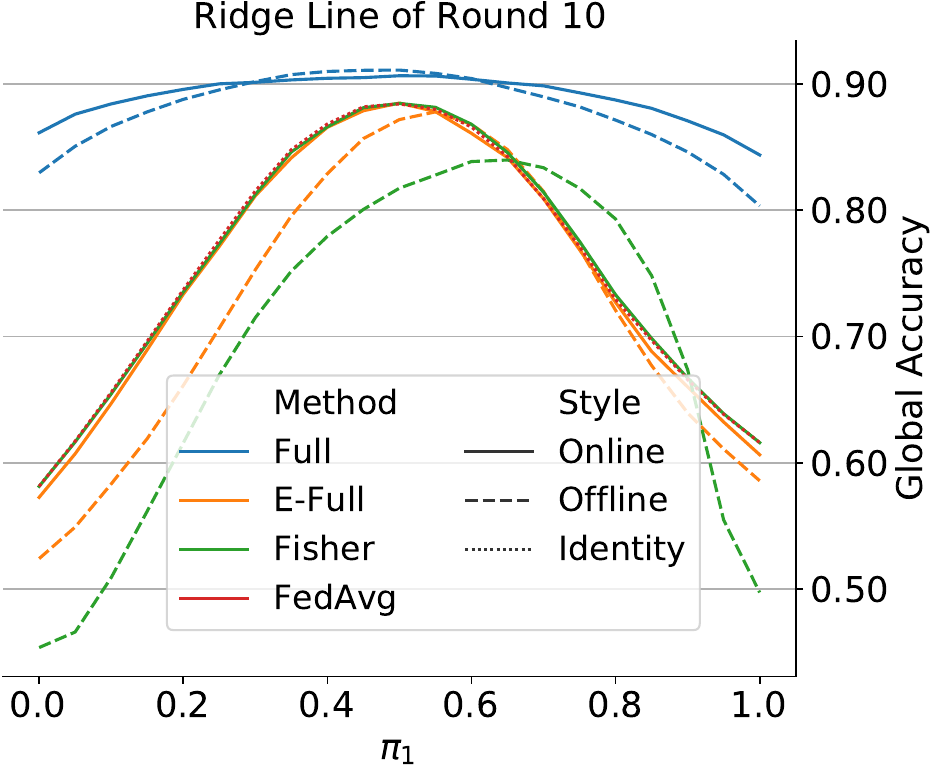}
        \label{Ridge10}
    }
    
    \caption{(a) The covariance of the first 1000 parameters in a normalized $\Sigma_{1}^{-1}$ evaluated by the full Laplace approximation. For normalization, we let $\Sigma_{1}^{-1}$ subtract the minimum value and divide by the maximum value. 
    (b) The global accuracy of round 1 using the model obtained by the aggregation functions with respect to $\pi_{1}$. 
    (c) The global accuracy of round 10, accordingly. 
    Different colors indicate different methods and different lines represent different evaluation styles. 
    Note that identity evaluation is only used in FedAvg.}
    \label{fig.AnalyzeHessian}
\end{figure*}

In this section, we investigate the performance of our aggregation strategy using three Laplace approximations (full, e-full and diagonal Fisher) under two evaluation styles (offline and online).
The full Laplace approximation calculates the Hessian matrix directly to estimate the covariance.
Both the e-full and the diagonal Fisher methods drop the correlations between parameters and approximate the covariance using a diagonal matrix. 
The e-full approximation method calculates the eigenvalues of the full Hessian to approximate the covariance, while the diagonal Fisher model uses the diagonal Fisher introduced in Sec. \ref{SEC:Estimation}.
The investigation is conducted on the MNIST dataset using a small neural network with a 780-10 architecture.
The computational cost of calculating the second derivative of the small model is within the scope of modern computers, so it is possible to estimate the full Hessian of likelihoods.
Besides, we only use two clients, each of which has five categories of data with similar numbers of samples. 
There are no shared categories between the two clients, which is equivalent to the extremely heterogeneous data setting of $\alpha=0$.
The global model used in the next round is aggregated at $\pi_{1}=0.51$.
Additionally, we select a learning rate $lr=0.01$, epoch $E=1$ and batch size $B=200$, which enables the local models to achieve local optima in the first round.
For comparison, we train two local models until they reach the optimal solution and aggregate them using Eq. \ref{EQ:ProductGaussian}, in which the covariance is evaluated by different Laplace approximations.
Specifically, we take $\pi_{i}$ as a variable, where $\pi_{i} \in [0,1]$ and $\sum_{i}^{N} \pi_{i} = 1$.
Therefore, the aggregation equation of our method is a univariate function $\theta_{S} = \pi_{1} \Sigma_{S} \Sigma^{-1}_{1} \mu_{1} + (1 - \pi_{1}) \Sigma_{S} \Sigma^{-1}_{2} \mu_{2}$, where $ \Sigma_{S}^{-1} = \pi_{1} \Sigma_{1}^{-1} + (1 - \pi_{1}) \Sigma_{2}^{-1}$ and $\pi_{1} \in [0,1]$.
As discussed in \cite{ray2005topography}, the above function of $\theta_{S}$ is the ridge line of the mixture density of posteriors. 
Correspondingly, the aggregation equation of FedAvg is a weighted sum function $\theta_{S} = \pi_{1}\mu_{1} + (1- \pi_{1}) \mu_{2}$.
In Fig. \ref{Ridge1} and \ref{Ridge10}, we draw the accuracy curves evaluated on the test dataset of MNIST using the above functions of $\theta_{S}$.

\begin{figure*}[t]
    \centering
        \subfigure[The cosine similarity between normalized $\Sigma_{1}^{-1}$ and $\Sigma_{2}^{-1}$ over 10 rounds.]{
        \includegraphics[width=2.2in]{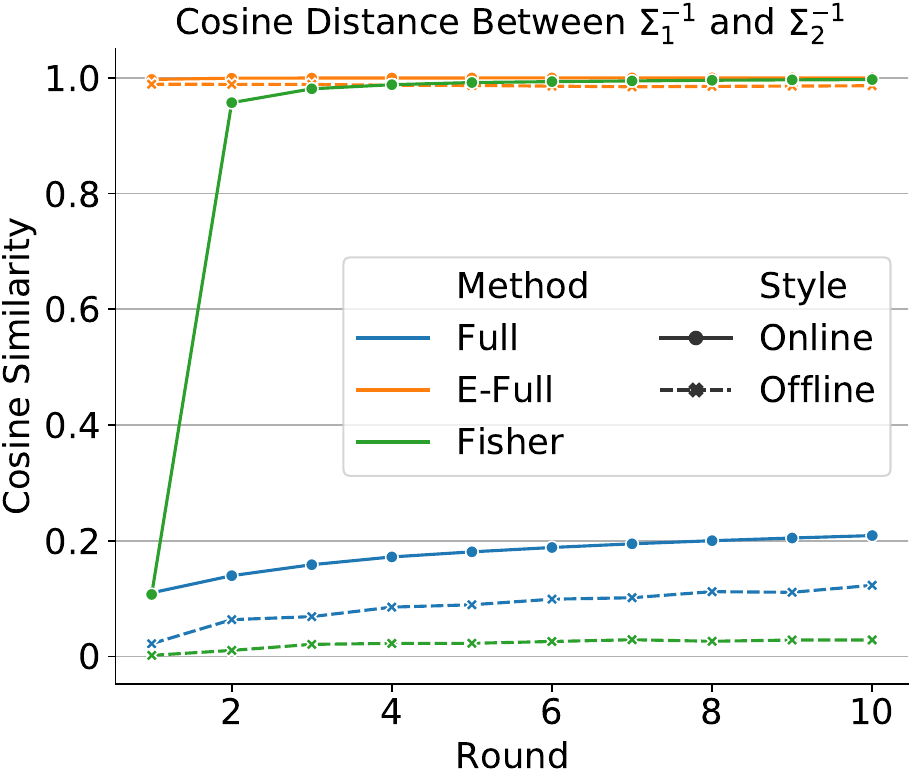}
        \label{CosDis}
    }
    \subfigure[]{
        \includegraphics[width=2.2in]{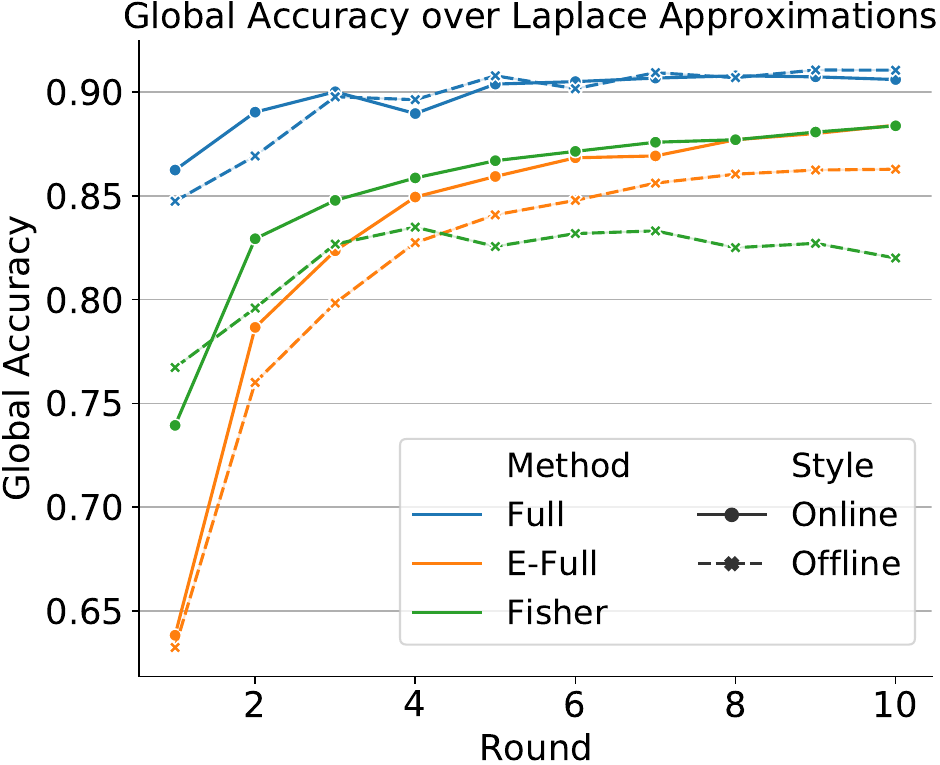}
        \label{R10GA}
    }
    \subfigure[]{
	\includegraphics[width=2.2in]{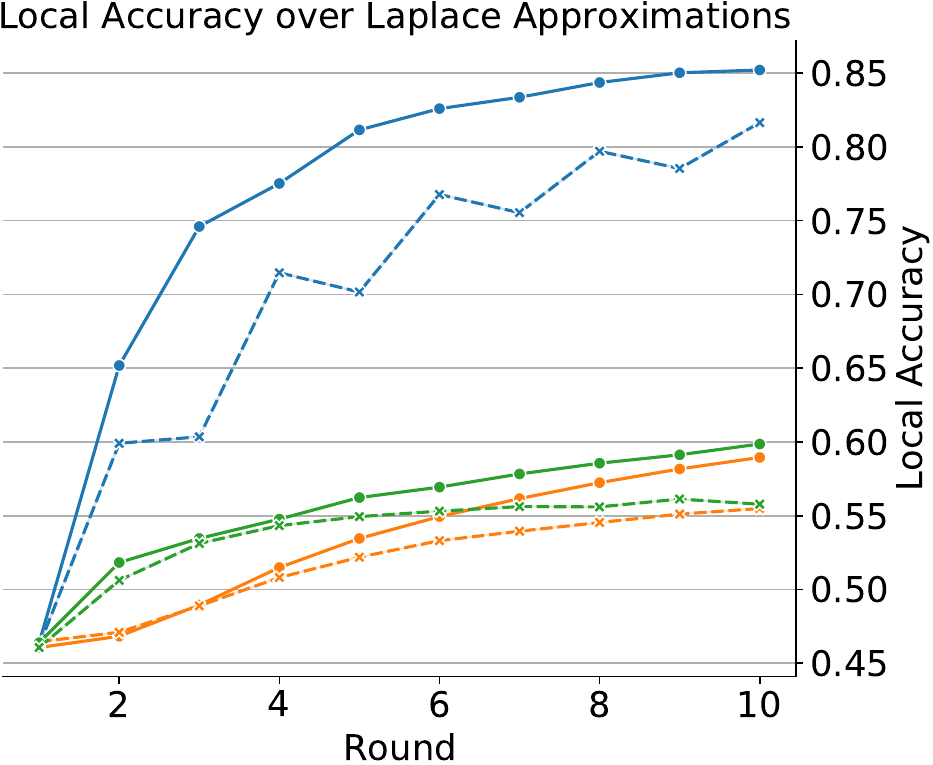}
        \label{R10LA}
    }
    
    \caption{(a) The cosine similarity between normalized $\Sigma_{1}^{-1}$ and $\Sigma_{2}^{-1}$ obtained by different Laplace approximations over different evaluation styles. We vectorize the matrix and directly calculate the cosine distance.
    (b) The global accuracy curves of 10 rounds. 
    (c) The local accuracy curves of 10 rounds. 
    Different colors indicate different methods and different lines represent different evaluation styles.}
    \label{fig.R10}
\end{figure*}

\textbf{Effect of evaluation styles.}
We compare the results obtained by different evaluation styles across different Laplace approximations. 
For the full approximation method, the difference in global accuracy between the online and offline evaluation is very small.
However, the online method obtains more stable local accuracy and better final local accuracy than the offline strategy as shown in \ref{R10LA}.
This means that the joint modes obtained online are better than those obtained offline.
For the diagonal Fisher methods, the global accuracy of the online method is lower than that of the offline one at first, but the final result is the opposite.
As shown in \ref{R10GA} and \ref{Ridge10}, the offline evaluation merely achieves a global accuracy of less than $83\%$ after 10 rounds, which is worse than the results of FedAvg. 

As for the e-full approximation, a similar phenomenon to the diagonal Fisher methods can be observed.
The offline evaluation produces worse accuracy curves, and both the final global and local accuracy are lower, compared with the online method. 

In general, from Fig. \ref{R10GA} and \ref{R10LA}, we can see that the online evaluation performs better than the offline method. 
The results of the online model are stable and convergent, while both the final global and local accuracy of the offline method are poor.

\begin{figure*}[t]
    \centering
        \subfigure[]{
        \includegraphics[width=3.0in]{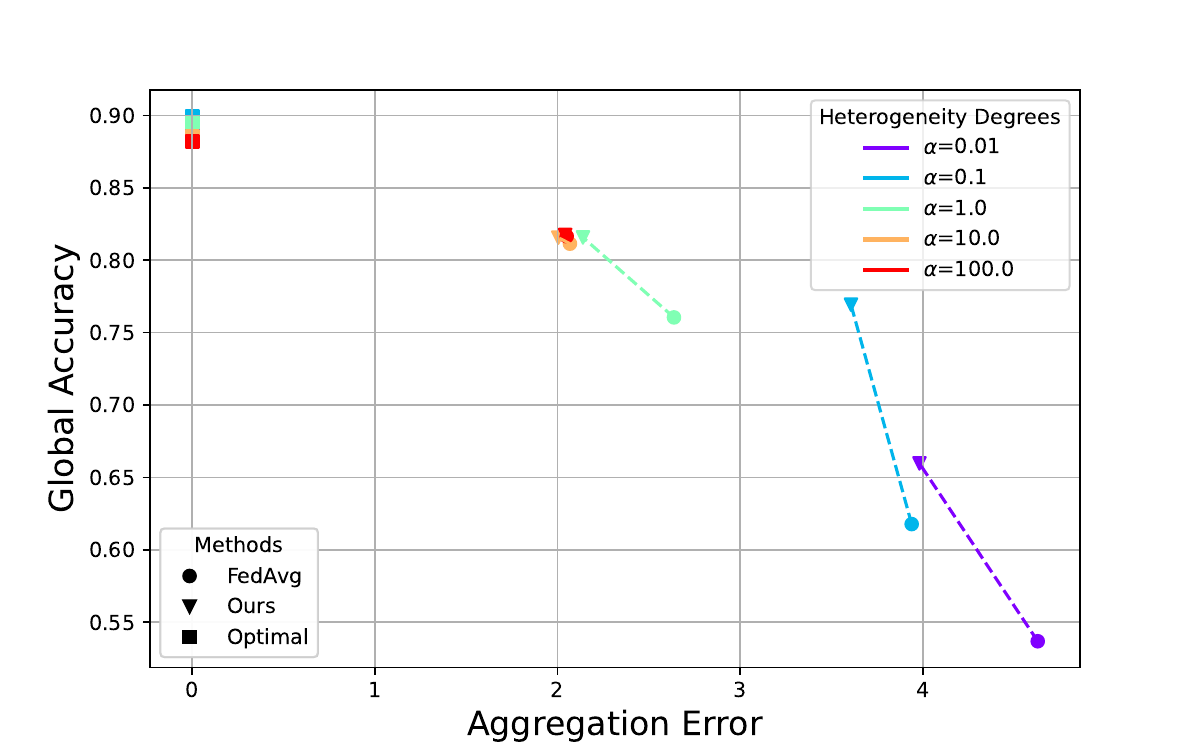}
        \label{GA}
    }
    \subfigure[]{
        \includegraphics[width=3.0in]{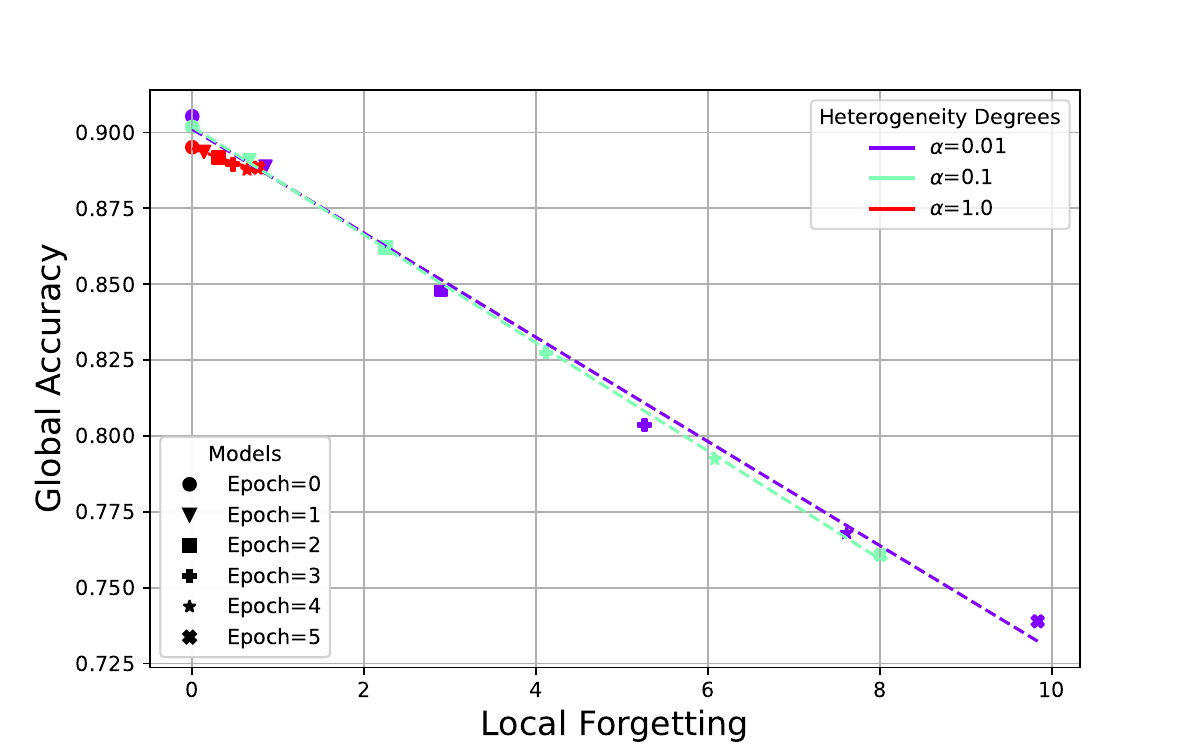}
        \label{LA}
    }
    \caption{(a) Correlation between GA and AE. `Optimal' refers to the model obtained through 15 rounds of full Laplace approximation, while our method and FedAvg are trained with 3 rounds. 
    (b) Correlation between LA and LF. Models with varying degrees of forgetting are obtained by setting different epoch counts.
    Different colors indicate different heterogeneity degrees of the federated data, and different shapes represent different models.}
    \label{fig.correlation}
\end{figure*}

\textbf{Correlations between Measures and Problems.} GA and LA are the two main metrics in this paper, which not only reflect the performance of the model but are also used to assess AE and LF, respectively. In this section, we will demonstrate the relevance of these metrics to these two issues. To quantify AE and LF, we assume the global posterior is known, represented as 
\( p(\theta^{*}|D) = \mathcal{N}(\mu^{*}_S, \Sigma^{*}_S) \). Given any model, the distance to the optimal model can be defined as the negative logarithm on its posterior density: 
\[ D(\theta, \theta^{*}) = -\ln(p(\theta|\theta^{*})) = \frac{1}{2} (\theta -\mu_{S}^{*})^{\top} \Sigma_{S}^{*{-1}} (\theta -\mu_{S}^{*}). \]
Thus, AE represents the probability density of the server-aggregated model, and LF represents the probability density of the model after local training. In actual experiments, we employ the full Laplace approximation to obtain the global posterior, leading us to the optimal model and its corresponding full Hessian matrix. We then demonstrate the relationship between GA and AE through different aggregation methods and the relationship between LA and LF through different local training epochs. The experiments are conducted on the MNIST dataset, with the number of clients set to 2. The experimental results are illustrated in the Fig.~\ref{fig.correlation}. The measure for AE is \(D(\theta_S, \theta^{*})\), where \(\theta_S\) is the aggregated model in the server and \(\theta^{*}\) is the optimal model. The aggregated model is taken from the third round of the training process, while the optimal model is taken from the 15th round after convergence. The LF is measured as \(\sum_{n} \pi_{n} D(\theta_n, \theta^{*})\), where \(\theta_n\) is trained model in the client taken from the 16th round. 
We tested the correlation under different heterogeneity degrees. When \(\alpha\) is relatively large, there's a more pronounced correlation between the metric and the problem. However, when \(\alpha\) is smaller, due to the reduced difficulty of federated learning, the correlation also weakens.

\textbf{Effect of the correlations.}
In this part, we mainly aim to study the effect of the correlations between parameters.
Although diagonal methods have been successfully and effectively used in many works, they ignore the correlation between parameters in the covariance, which is also an important factor for reducing AE. 
As shown in Fig. \ref{FullHessian}, there is unmissable covariance in a full Hessian but the diagonal methods only use the variance on the diagonal. 
If some parameters exhibit high covariance, ignoring the correlations will turn the high-probability areas of the true posterior density into low-probability areas, weakening the ability of our aggregation strategy to reduce AE.
In contrast, the full Laplace approximation incurs massive computational costs but effectively estimates the correlations between parameters.

As shown in Fig. \ref{Ridge1}, \ref{Ridge10} and \ref{R10GA}, whether using offline or online evaluation, the full approximation method achieves the best results.
Further, this only requires a few rounds of training to converge.
In fact, it takes only 3 rounds time to achieve nearly $90\%$ global accuracy, while other methods need more than 10 rounds.
Besides, the curve areas of the full approximation methods in Fig. \ref{Ridge1} and \ref{Ridge10} are also larger than others.
Local models obtained by the full approximation method achieve more than $80\%$ local accuracy after 10 rounds while these of other methods only achieve less than $60\%$ local accuracy.
Those observations suggest that the full approximation methods are much better at finding joint modes than other methods. 
 
As for the e-full approximation methods, they calculate the eigenvalues of the full Hessian, which means their covariance contains key information of the full Hessian.
However, they still suffer from lacking of the correlations, leading to much lower global and local accuracy than the full approximation method.
In fact, they perform similar to FedAvg. 
As shown in \ref{CosDis}, the matrices $\Sigma_{1}^{-1}$ and $\Sigma_{2}^{-1}$ obtained by the e-full approximation methods are very similar, so their effect is almost equivalent to that of the identity array for aggregation.
However, because some elements have different values, the results of e-full approximation methods will be slightly different from those of FedAvg.

In contrast, the covariance obtained by the diagonal Fisher methods can retain the difference.
Compared with the e-full approximation methods, this difference allows our method to improve the accuracy of the aggregated model in the first few rounds.
Therefore, calculating the correlations is an effective way to boost the results of our framework.
However, the full approximation methods require high computational and storage overhead. 
If we consider efficiency, the best method for our framework is the diagonal Fisher method with offline evaluated.


\section{Conclusion\label{SEC:Con}}

Motivated by the Bayesian and probability theories, we propose a novel FL framework, which includes a new aggregation strategy on the server side and a new training strategy on the client side. 
This new FL framework can improve the accuracy of aggregated models and the generalization ability of client models. 
Based on the assumption that the posterior probability follows a Gaussian distribution, our aggregation strategy treats the process of parameter aggregation as a product of Gaussians. Thus we can easily estimate the joint modes using the means and covariances of the distribution. 
In addition, when training the client models, we develop a prior loss from the distributed posterior probabilistic parameters.
The regularization term can help maintain the generalization ability of the local models.
Moreover, our method can improve the convergence rate with parameter compression.
Experiments results clearly show that our method achieves state-of-the-art results on commonly used FL benchmarks. 
In the future, it would be interesting to explore other covariance approximation methods such as the block-diagonal of the Fisher \cite{martens2016second,ritter2018online}, to approximate the posteriors.



\ifCLASSOPTIONcaptionsoff
  \newpage
\fi


\bibliographystyle{IEEEtran}
\bibliography{references.bib}

%








\begin{IEEEbiography}[{\includegraphics[width=1in,height=1.25in,clip,keepaspectratio]{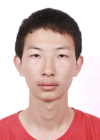}}]{Liangxi Liu} received the B.S. degrees from Department of Computer Science and Engineering at Southern University of Science and Technology, Shenzhen, China, 2019. Currently, he is a Research Assistant in SUSTech, Shenzhen, China. His current research interests include machine learning and computer vision. 
\end{IEEEbiography}
\vspace{-1cm}
\begin{IEEEbiography}[{\includegraphics[width=1in,height=1.25in,clip,keepaspectratio]{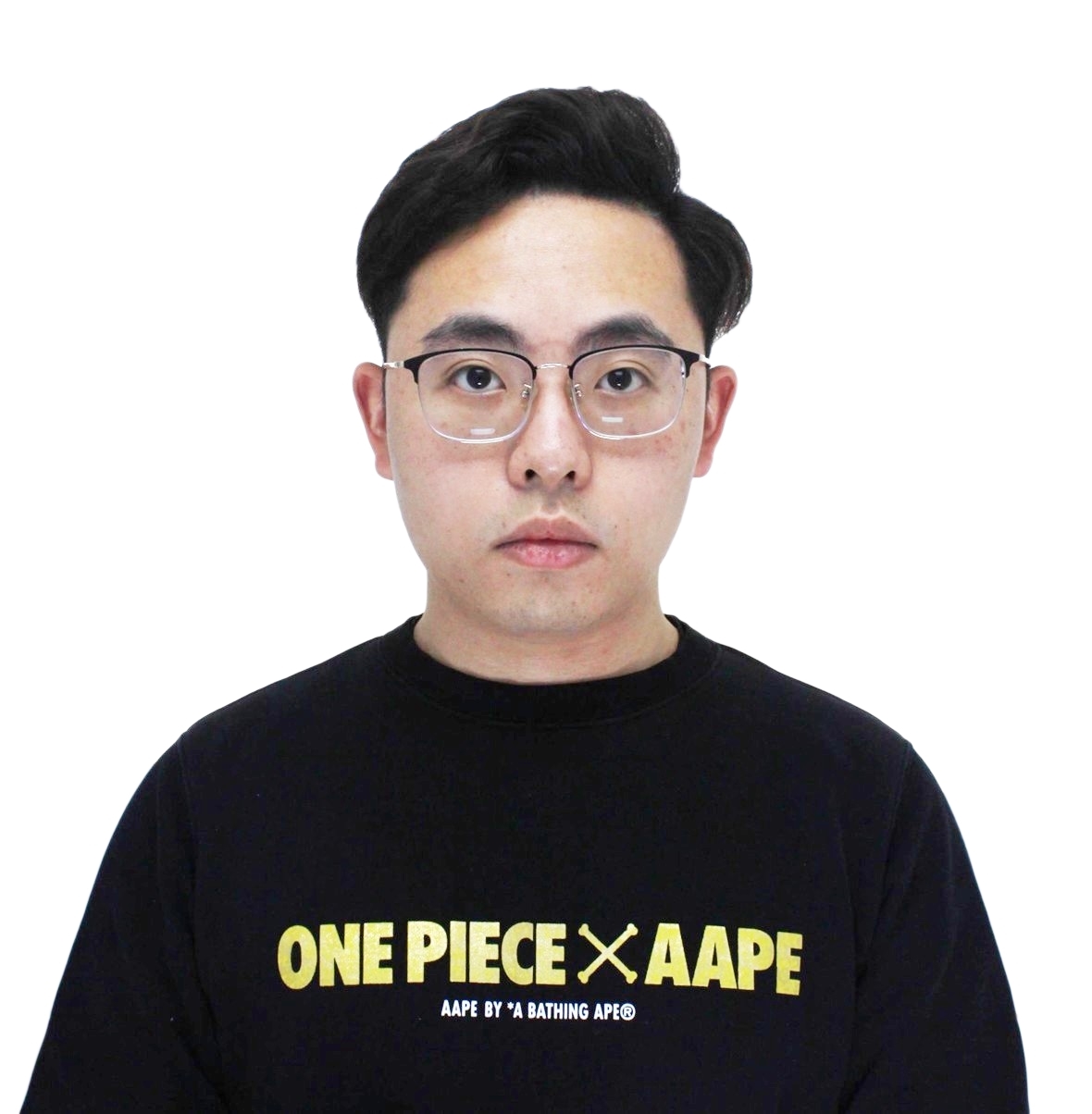}}]{Xi Jiang} received his B.S. degree from Xi'an Jiaotong University, China, in 2020 and his M.S. degree from the Southern University of Science and Technology, China, in 2023. He is currently a Ph.D. candidate in the Department of Computer Science and Engineering at Southern University of Science and Technology. His research interests include computer vision and machine learning.
\end{IEEEbiography}
\vspace{-1cm}
\begin{IEEEbiography}[{\includegraphics[width=1in,height=1.25in,clip,keepaspectratio]{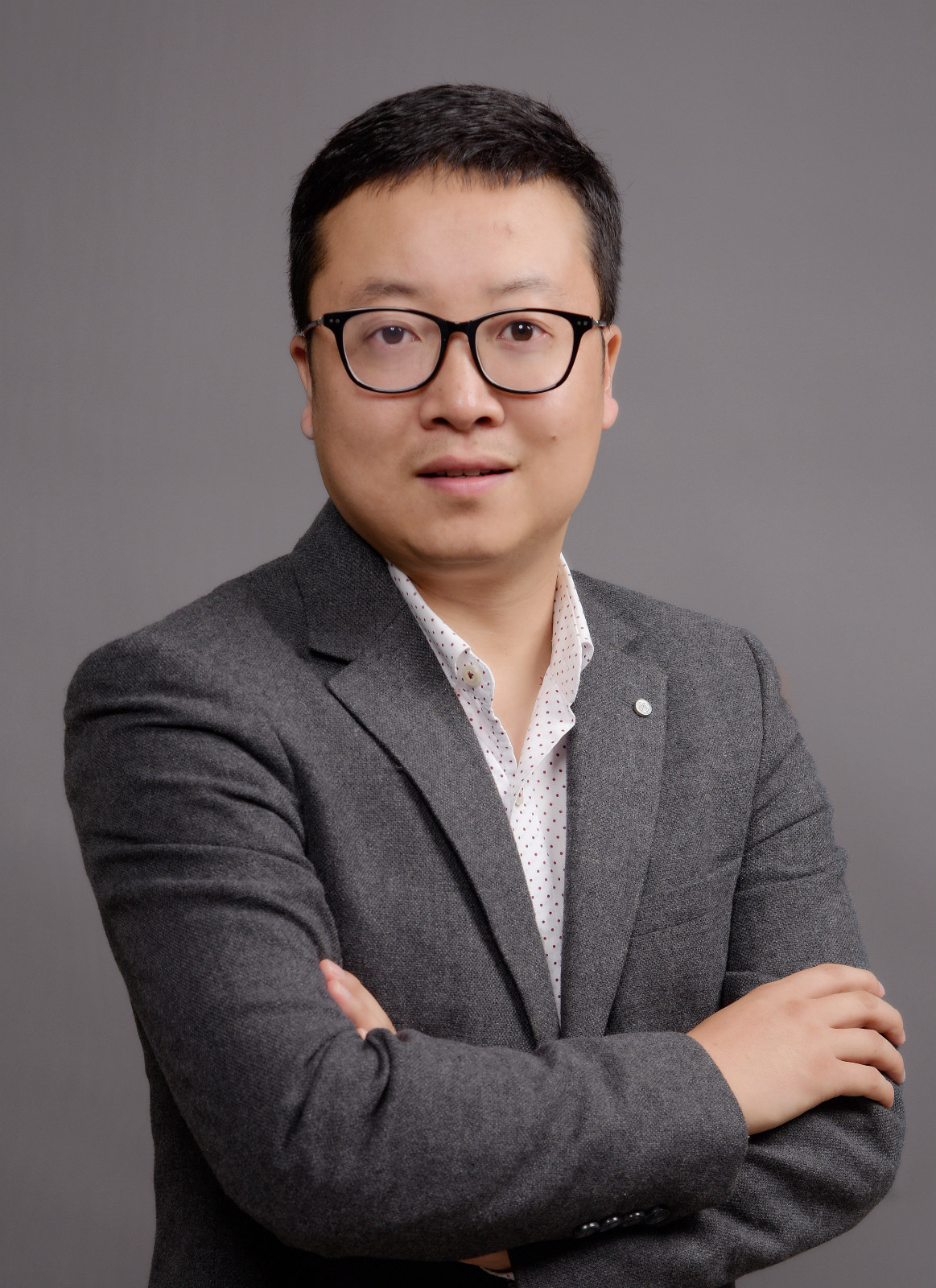}}]{Feng Zheng (M'19)} received Ph.D. degree from The University of Sheffield, UK, 2017. He is currently the Associate Professor in Department of Computer Science and Engineering at Southern University of Science and Technology, Shenzhen, China. His research interests include machine learning, computer vision and cross-media intelligence.
\end{IEEEbiography}
\vspace{-1cm}
\begin{IEEEbiography}[{\includegraphics[width=1in,height=1.25in,clip,keepaspectratio]{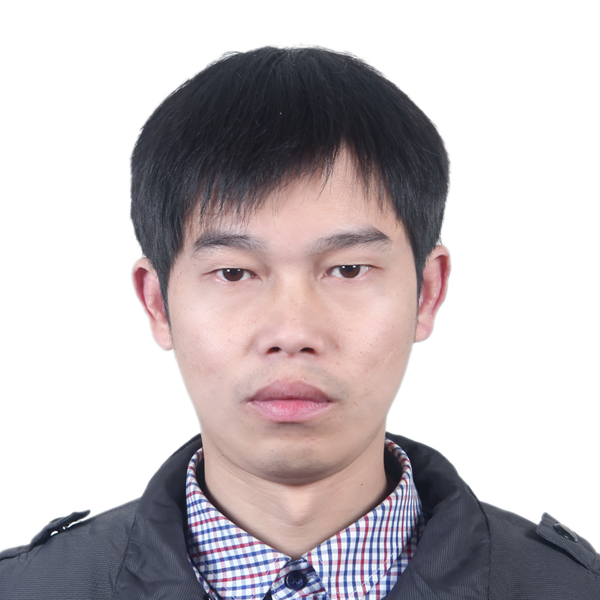}}]{Hong Chen} received the B.S., M.S. and Ph.D. degrees from Hubei University, Wuhan, China, in 2003, 2006, and 2009, respectively. Currently, he is a professor in the Department of Mathematics and Statistics, College of Science, Huazhong Agricultural University, Wuhan, China. His current research interests include machine learning, statistical learning theory, and approximation theory. 
\end{IEEEbiography}
\vspace{-1cm}
\begin{IEEEbiography}[{\includegraphics[width=1in,height=1.25in,clip,keepaspectratio]{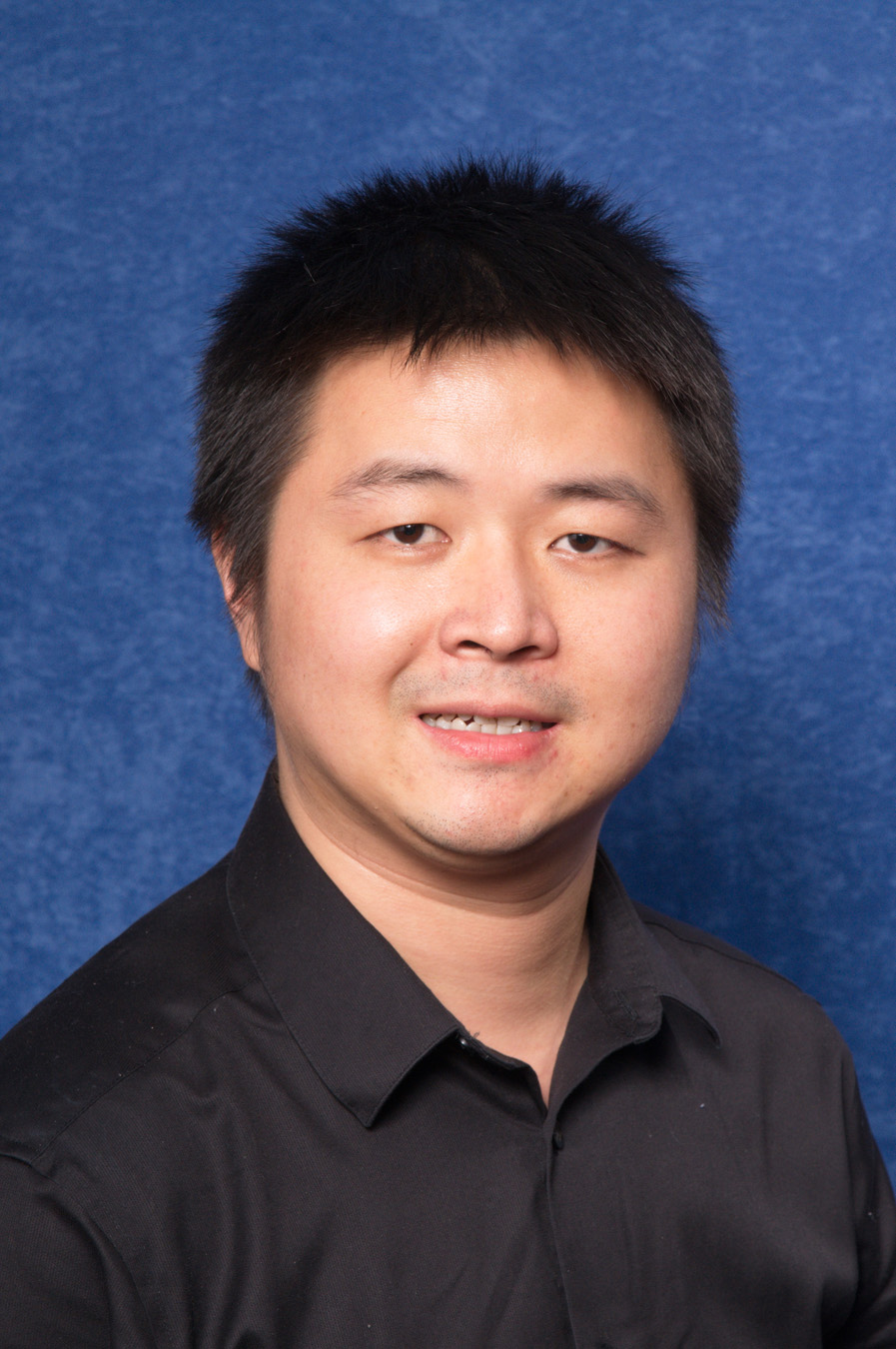}}]{Guo-Jun Qi (Fellow’21)} is a Professor at Westlake University and the Chief Scientist leading and overseeing  Artificial Intelligence Research at OPPO. Prior to that, He was the Chief AI Scientist at Futurewei, a faculty member in the Department of Computer Science at the University of Central Florida since August 2014. Dr. Qi has published over 150 papers in a broad range of venues. Among them are the best student paper of ICDM 2014, ``the best ICDE 2013 paper" by IEEE Transactions on Knowledge and Data Engineering, as well as the best paper (finalist) of ACM Multimedia 2007 (2015).
\end{IEEEbiography}
\vspace{-1cm}
\begin{IEEEbiography}[{\includegraphics[width=1in,height=1.25in,clip,keepaspectratio]{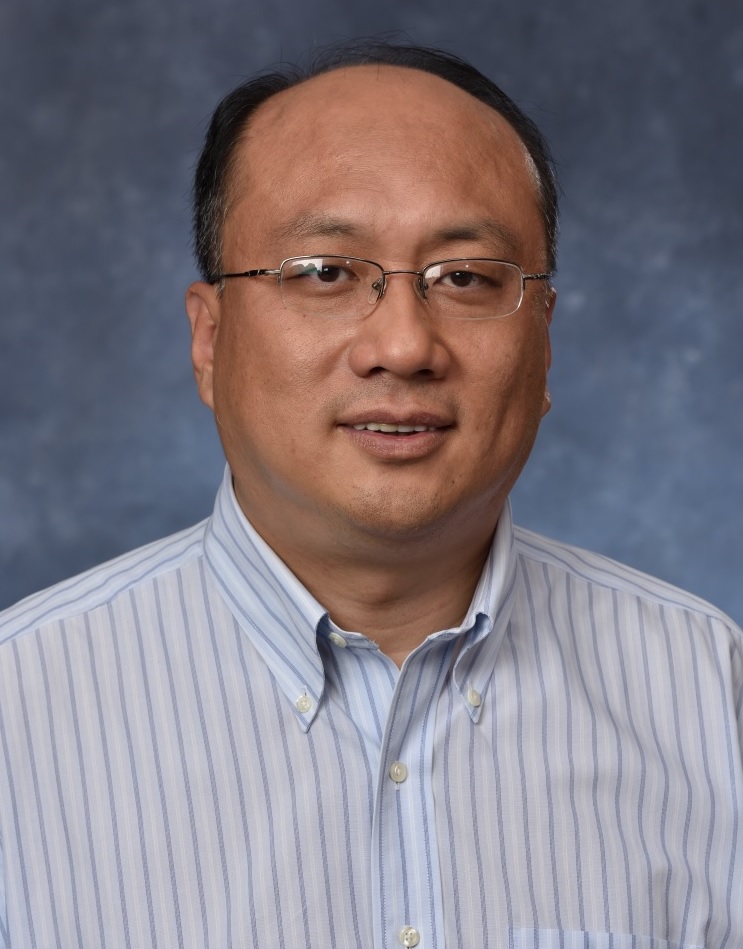}}]{Heng Huang} received both B.S. and M.S. degrees from Shanghai Jiao Tong University, China, in 1997 and 2001, respectively. He received the Ph.D. degree in Computer Science from Dartmouth College in 2006. Currently, he is a Brendan Iribe Endowed Professor in computer science with the University of Maryland College Park, College Park, MD, USA. His research interests include machine learning, data mining, biomedical data science, bioinformatics, and neuroinformatics.
\end{IEEEbiography}
\vspace{-1cm}
\begin{IEEEbiography}[{\includegraphics[width=1in,height=1.25in,clip,keepaspectratio]{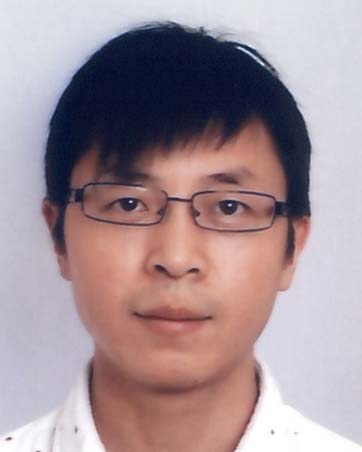}}]{Ling Shao (Fellow’20)} is a Distinguished Professor with the UCAS-Terminus AI Lab, University of Chinese Academy of Sciences, Beijing, China. He was the founding CEO and Chief Scientist of the Inception Institute of Artificial Intelligence, Abu Dhabi, UAE. He was also the Initiator, founding Provost and EVP of MBZUAI, UAE. His research interests include generative AI, vision and language, and AI for healthcare. He is a fellow of the IEEE, the IAPR, the BCS and the IET.
\end{IEEEbiography}
\vspace{15cm}

\end{document}


%
\title{Appendix of A Bayesian Federated Learning Framework with Online Laplace Approximation}

\markboth{}%
{}
%

\author{Liangxi Liu, Feng Zheng$^*$~\IEEEmembership{Member,~IEEE}, Hong Chen, Guo-Jun Qi~\IEEEmembership{Senior Member,~IEEE}, Heng Huang, Ling Shao~\IEEEmembership{Fellow,~IEEE}   %
    \thanks{Manuscript received XXX XX, 2021. ($^*$\emph{Corresponding author: Feng Zheng})}%
    \thanks{L. Liu and F. Zheng are with Southern University of Science and Technology, Shenzhen 518055, China (email: liulx@mail.sustech.edu.cn, and f.zheng@ieee.org).}
    \thanks{H. Chen is with Huazhong Agricultural University, Wuhan, China. (email: chenh@mail.hzau.edu.cn)}
    \thanks{G. Qi is with the Futurewei Technologies Seattle WA, USA. (e-mail: guojunq@gmail.com)}
    \thanks{H. Huang is with University of Pittsburgh, Pittsburgh, PA, USA. (email: heng.huang@pitt.edu)}
    \thanks{L. Shao is with Inception Institute of Artificial Intelligence, UAE. (email: ling.shao@ieee.org)}

    \thanks{Code is available at \href{https://github.com/Klitter/A-Bayesian-Federated-Learning-Framework-with-Online-Laplace-Approximation}{{\color{black}https://github.com/Klitter/A-Bayesian-Federated-Learning-Framework-with-Online-Laplace-Approximation}}}
    }



\maketitle

%
\IEEEpeerreviewmaketitle

\appendices

\section{A simple proof for the Multivariate Gaussian Mixture Theory from \cite{ray2005topography}}

Here, we try to prove that the critical points of two Gaussian posterior distributions $p\left(\theta | \mathcal{D}_{1}\right)$ and $p\left(\theta | \mathcal{D}_{2}\right)$ lie on a hypersurface $\theta=\left(\alpha_{1} \Sigma_{1}^{-1}+\alpha_{2} \Sigma_{2}^{-1}\right)^{-1}\left(\alpha_{1} \Sigma_{1}^{-1} \mu_{1}+\alpha_{2} \Sigma_{2}^{-1} \mu_{2}\right)$, where $ \alpha_{1} \in[0,1]$ and $\alpha_{1}+\alpha_{2}=1$:

\begin{equation}
p\left(\theta | \mathcal{D}_{1}\right) \approx \mathcal{N}\left(\theta | \mu_{1}, \Sigma_{1}\right), 
p\left(\theta | \mathcal{D}_{2}\right) \approx \mathcal{N}\left(\theta | \mu_{2}, \Sigma_{2}\right).
\end{equation}

The derivative of all critical points on the density of the mixture of $p\left(\theta | \mathcal{D}_{1}\right)$ and $p\left(\theta | \mathcal{D}_{2}\right)$ is equal to zero:
\begin{equation}
\frac{\partial\left(p\left(\theta | \mathcal{D}_{1}\right) +p\left(\theta | \mathcal{D}_{2}\right)\right)}{\partial \theta}=0.
\label{mgm}
\end{equation}

The expression of the critical point can be simply obtained by the above formula with $\alpha_{1}$ = $\frac{p\left(\theta | \mathcal{D}_{1}\right)}{p\left(\theta | \mathcal{D}_{1}\right)+p\left(\theta | \mathcal{D}_{2}\right)} \in [0,1]$ and $\alpha_{2}=1-\alpha_{1}$:

\begin{equation}
\theta=\left(\alpha_{1} \Sigma_{1}^{-1}+\alpha_{2} \Sigma_{2}^{-1}\right)^{-1}\left(\alpha_{1} \Sigma_{1}^{-1} \mu_{1}+\alpha_{2} \Sigma_{2}^{-1} \mu_{2}\right)
\end{equation}

Therefore, our claim is true.

\section{Federated Online Laplace approximation on the Global Posterior $p(\theta \mid \mathcal{D}) $}
In this section, we directly derive the formula of the global covariance by applying federated online Laplace approximation when approximating the global posteriors $p(\theta \mid \mathcal{D})$.

In round $r=1$, the global shared model is initialized and distributed to each client as a prior for the local training. At this time, without any data information, the expectation is a zero matrix $O$ and the inverse of the covariance of the global posterior is set to $\gamma I$, where $\gamma$ is a small number such as $10^{-4}$. Therefore, the prior for the first round of local training is $p_{r=1}(\theta) \approx \mathcal N(O, \gamma I)$.

At round $r=R$, the global posterior $p_{R}(\theta \mid \mathcal{D})$ can be disassembled into three parts: the likelihood $p_{R} \left( \mathcal{D}|\theta \right)$, the prior $ p_{R}(\theta)$ and the data distribution $\ln p_{R}(\mathcal{D})$:
\begin{equation}
    \ln p_{R}(\theta \mid \mathcal{D}) = \ln p_{R} \left( \mathcal{D}|\theta \right) + \ln p_{R}(\theta) - \ln p_{R}(\mathcal{D}) .
\end{equation}
Generally, we can assume that the local datasets $ \mathcal{D}_{n}$ for each client are independent of each other, so we can decompose the global likelihood into the product of local likelihoods $p_{R} \left( \mathcal{D}|\theta \right) = \prod_{n} p_{R} \left( \mathcal{D}_{n}|\theta \right)$:

\begin{equation}
\ln p_{R}(\theta \mid \mathcal{D})  =  \sum_{n} \ln p_{R}\left( \mathcal{D}_{n} |\theta \right) + \ln p_{R}(\theta) - \ln p_{R}(\mathcal{D}) .
\label{eq.gpll}
\end{equation}

Then, we apply prior iteration (PI) to regard the prior as the global posterior of the last round $ p_{R}(\theta) = p_{R-1}(\theta | \mathcal{D})$:
\begin{equation}
    \ln p_{R}(\theta \mid \mathcal{D})  =  \sum_{n} \ln p_{R}\left( \mathcal{D}_{n} |\theta \right) + \ln p_{R-1}(\theta | \mathcal{D}) - \ln p_{R}(\mathcal{D}).
\end{equation}
Similar to Eq. \ref{eq.gpll}, we decompose the global posterior of the last round:
\begin{equation}
\begin{split}
    \ln p_{R}(\theta \mid \mathcal{D})  &=  \sum_{n} \ln p_{R}\left( \mathcal{D}_{n} |\theta \right) + \ln p_{R-1}(\theta | \mathcal{D}) - \ln p_{R}(\mathcal{D}) \\
    &=  \sum_{r=R-1}^{R} \sum_{n} \ln p_{r}\left( \mathcal{D}_{n} |\theta \right) + \ln p_{R-1}(\theta) \\
    & \hspace{4cm} - \sum_{r=R-1}^{R} \ln p_{r}(\mathcal{D}).
\end{split}
\end{equation}

If we apply PI to all the previous rounds, we will obtain a global posterior integrating all local and previous likelihoods:  
\begin{equation}
\begin{split}
    \ln p_{R}(\theta \mid \mathcal{D}) =  \sum_{r=1}^{R} \sum_{n} \ln p_{r}\left( \mathcal{D}_{n} |\theta \right) &+ \ln p_{r=1}(\theta) \\
    & \hspace{0.5cm} - \sum_{r=1}^{R} \ln p_{r}(\mathcal{D}) .
\label{eq.gp}
\end{split}
\end{equation}

By applying the Laplace approximation to Eq. \ref{eq.gp}, we derive the formula of the global covariance:

\begin{align}
    \Sigma^{-1}_{S,R} &= \bar{F}_{S,R} = \frac{1}{R} \mathbb{E}_{(x,y) \in \mathcal{D}} [\nabla^{2} \ln p_{R}(\theta \mid \mathcal{D}) ] \nonumber  \\
    &=  \frac{1}{R \cdot | \mathcal{D}|} \sum_{r=1}^{R} \sum_{n} \nabla^{2} \ln p_{r}\left( \mathcal{D}_{n} |\theta \right) + \nabla^{2} \ln p_{r=1}(\theta) \nonumber \\
    &= \frac{1}{R} \sum_{r=1}^{R} \sum_{n} \pi_{n} \bar{F}_{n,r} +  \gamma I  \label{eq.gcol}  \\ 
    &= \frac{1}{R} \sum_{r=R-1}^{R} \sum_{n} \pi_{n} \bar{F}_{n,r} + \frac{R-1}{R} \sum_{n} \pi_{n} \Sigma_{n,R-1} \nonumber \\
    &= \sum_{n} \pi_{n} \Sigma^{-1}_{n,R}. \label{eq.agggc}
\end{align}
Note that, Eq. \ref{eq.agggc} is our aggregation method, which obtains the global covariance by summing all the local covariances.

Additionally, Eq. \ref{eq.gcol} illustrates that the global covariance obtained by our method is a federated and online-evaluated item which integrates all local and past Fishers. Therefore, it is much more correct than the covariance obtained by offline methods, and can be directly used in the FL framework.

\ifCLASSOPTIONcaptionsoff
  \newpage
\fi


\bibliographystyle{IEEEtran}
\bibliography{references.bib}

%







